\documentclass{article}

\usepackage{microtype}
\usepackage{graphicx}
\usepackage{booktabs} 

\usepackage{hyperref}

\usepackage{amsmath}
\usepackage{amsthm}
\usepackage{amssymb}
\usepackage{newcommand}

\usepackage{caption}
\usepackage{subcaption}

\usepackage{adjustbox}

\usepackage{algorithm}
\usepackage{algorithmic}
\usepackage{wrapfig}

\usepackage{adjustbox}
\usepackage{hyperref}
\usepackage[normalem]{ulem}

\usepackage{tikz,tkz-graph}
\usetikzlibrary{
  graphs,
  graphs.standard,
  positioning,chains,fit,shapes,calc,
}
\usepackage{graphicx}
\usepackage{wrapfig}

\usepackage{amsmath} 
\usepackage{tikz}
\usepackage{etoolbox} 
\usepackage{listofitems} 
\usetikzlibrary{arrows.meta} 
\usepackage[outline]{contour} 
\contourlength{1.4pt}

\tikzset{>=latex} 
\usepackage{xcolor}
\colorlet{myred}{red!80!black}
\colorlet{myblue}{blue!80!black}
\colorlet{myyellow}{yellow!80!black}
\colorlet{mygreen}{green!60!black}
\colorlet{myorange}{orange!70!red!60!black}
\colorlet{mydarkred}{red!30!black}
\colorlet{mydarkblue}{blue!40!black}
\colorlet{mydarkgreen}{green!30!black}
\tikzstyle{node}=[very thick,circle,draw=myblue,minimum size=22,inner sep=0.5,outer sep=0.6]
\tikzstyle{node in}=[node,green!15!black,draw=mygreen,fill=mygreen!25]
\tikzstyle{node hidden}=[node,yellow!15!black,draw=myyellow,fill=myyellow!20]
\tikzstyle{node hidden2}=[node,blue!15!black,draw=myblue,fill=myblue!20]
\tikzstyle{node convol}=[node,orange!15!black,draw=brown,fill=brown!20]
\tikzstyle{node out}=[node,red!15!black,draw=myred,fill=myred!20]
\tikzstyle{connect}=[thick,mydarkblue] 
\tikzstyle{connect arrow}=[-{Latex[length=4,width=3.5]},thick,mydarkblue,shorten <=0.5,shorten >=1]
\tikzset{ 
  node 0/.style={node convol},
  node 1/.style={node in},
  node 2/.style={node hidden},
  node 3/.style={node out},
}

\newcommand{\name}{DDOM}

\newcommand{\sm}[1]{#1}
\newcommand\boldmathred[1]{\textcolor{violet}{\mathbf{#1}}}
\newcommand\boldmathblue[1]{\textcolor{blue}{\mathbf{#1}}}

\usepackage[accepted]{icml2023}

\usepackage{amsmath}
\usepackage{amssymb}
\usepackage{mathtools}
\usepackage{amsthm}

\usepackage[capitalize,noabbrev]{cleveref}

\theoremstyle{plain}

\theoremstyle{definition}

\theoremstyle{remark}

\usepackage[textsize=tiny]{todonotes}

\icmltitlerunning{Diffusion Models for Black-Box Optimization}

\begin{document}

\twocolumn[
\icmltitle{Diffusion Models for Black-Box Optimization}




\begin{icmlauthorlist}
\icmlauthor{Siddarth Krishnamoorthy}{ucla}
\icmlauthor{Satvik Mashkaria}{ucla}
\icmlauthor{Aditya Grover}{ucla}
\end{icmlauthorlist}

\icmlaffiliation{ucla}{Department of Computer Science, UCLA}

\icmlcorrespondingauthor{Siddarth Krishnamoorthy}{siddarthk@cs.ucla.edu}

\icmlkeywords{Machine Learning, ICML}

\vskip 0.3in
]



\printAffiliationsAndNotice{}  

\begin{abstract}
 The goal of offline black-box optimization (BBO) is to optimize an expensive black-box function using a fixed dataset of function evaluations.
 Prior works consider \textit{forward} approaches that learn surrogates to the black-box function and \textit{inverse} approaches that directly map function values to corresponding points in the input domain of the black-box function.
 These approaches are limited by the quality of the offline dataset and the difficulty in learning one-to-many mappings in high dimensions, respectively. 
 We propose Denoising Diffusion Optimization Models (\name{}), a new inverse approach for offline black-box optimization based on diffusion models. 
 Given an offline dataset, \name{} learns a conditional generative model over the domain of the black-box function conditioned on the function values.
 We investigate several design choices in \name{}, such as reweighting the dataset to focus on high function values and  the use of classifier-free guidance at test-time to enable generalization  to function values that can even exceed the dataset maxima.
 Empirically, we conduct experiments on the Design-Bench benchmark~\citep{design-bench} and show that \name{} achieves results competitive with state-of-the-art baselines.
Our implementation of \name{} can be found at \url{https://github.com/siddarthk97/ddom}.
\end{abstract}

\section{Introduction}
Many fundamental problems in science and engineering involve optimization of an expensive black-box function, such as optimal experimental design and product optimization.
Since evaluating the black-box function is expensive, recent works consider purely data-driven approaches that use \textit{only} offline logged datasets to optimize the target function sidestepping real-world interactions~\citep{coms,mins,hansen2016cma}. 
We refer to this paradigm as offline black-box optimization (BBO).

\begin{figure*}[ht!]
\begin{subfigure}{\textwidth}
\begin{adjustbox}{width=\textwidth}
\begin{tikzpicture}
\def\xu{-6}
\def\yv{-1}
\def\xa{-3}
\def\ya{-1}
\def\xb{\xa+6}
\def\xc{\xb+2+1.5}
\def\yb{\ya+3.5}
\def\yq{\ya + 0.5}
\def\yz{\yq-1.75}

\node[rectangle, draw, rounded corners, minimum width=0.5cm, minimum height=1.5cm, fill=mygreen!10] (fd1) at (\xu,\yv+3) {\tiny $\mathbf{x}_0$};
\node[rectangle, draw, rounded corners, minimum width=0.5cm, minimum height=1.5cm, fill=mygreen!10] (fd2) at (\xu+1,\yv+3) {\tiny $\mathbf{x}_1$};
\node[rectangle, draw, rounded corners, minimum width=0.5cm, minimum height=0.5cm, fill=myred!10] (condy) at (\xu+6,\yv+4.5) {$y$};
\node[rectangle, draw, rounded corners, minimum width=0.5cm, minimum height=0.5cm, fill=myred!10] (condy2) at (\xu+1,\yv+4.5) {$y$};
\node[rectangle, draw, rounded corners, minimum width=0.5cm, minimum height=0.5cm, fill=myred!10] (condy3) at (\xu,\yv+4.5) {$y$};
\node[rectangle, draw, rounded corners, minimum width=0.5cm, minimum height=0.5cm, fill=myred!10] (condy4) at (\xu+2.75,\yv+4.5) {$y$};
\node[inner sep=0.pt] (nn) at (\xu+4.5,\yv+3) {\small $\nabla_\mathbf{x} \log p_t(\mathbf{x} | y)$};
\node[rectangle, draw, dashed, rounded corners, minimum width=2cm, minimum height=1cm] (nnbox) at (\xu+4.5,\yv+3) {};
\node[black] at (\xu+4.5, \yv+2.25) {\tiny Score function $\epsilon_\theta$};

\filldraw[black] (\xu+2,\yv+3) circle (2pt);
\filldraw[black] (\xu+1.75,\yv+3) circle (2pt);

\node[rectangle, draw, dashed, rounded corners, minimum width=7cm, minimum height=2.25cm] (revdif) at (\xu+3,\yv+3) {};
\node[black] at (\xu+3, \yv+2) {\small};

\node[rectangle, draw, rounded corners, minimum width=0.5cm, minimum height=1.5cm, fill=mygreen!10] (fdt2) at (\xu+2.75,\yv+3) {\tiny$\mathbf{x}_{\scalebox{0.75}{\textit{T-1}}}$};
\node[rectangle, draw, rounded corners, minimum width=0.5cm, minimum height=1.5cm, fill=mygreen!10] (fdt1) at (\xu+6,\yv+3) {\tiny $\mathbf{x}_T$};
\draw[thick, ->] (fd2.west) to[out=-180, in=0] node[right, align=center]{} (fd1.east);
\draw[thick, ->] (nnbox.west) to[out=-180, in=0] node[right, align=center]{} (fdt2.east);
\draw[thick, ->] (fdt1.west) to[out=-180, in=0] node[right, align=center]{} (nnbox.east);
\draw[thick, ->] (condy.west) to[out=-180, in=90] node[right, align=center]{} (nnbox.north);
\node[black] at (\xu+3, \yv+1.65) {\small Reverse diffusion};

\draw[thick, ->] (condy2.south) to[out=-90, in=90] node[right, align=center]{} (fd2.north);
\draw[thick, ->] (condy3.south) to[out=-90, in=90] node[right, align=center]{} (fd1.north);
\draw[thick, ->] (condy4.south) to[out=-90, in=90] node[right, align=center]{} (fdt2.north);

\node[rectangle, draw, dashed, rounded corners, minimum width=4cm, minimum height=2.25cm] (rwt) at (\xu-3.5+0.5,\yv+6) {};
\filldraw[color=red!60, fill=red!5, very thick](\xu-4.5+0.5,\yv+3.5+3) circle (0.2);
\filldraw[color=blue!60, fill=blue!5, very thick](\xu-2.5+0.5,\yv+3.5+3-0.15) circle (0.65);
\filldraw[color=green!60, fill=green!5, very thick](\xu-3.5+0.5-0.3,\yv+2.5+3) circle (0.35);
\node[black] at (\xu-3.5+0.5, \yv+7.25+0.10) {\small Reweighted $\mathcal{D}$};

\node[rectangle, draw, dashed, rounded corners, minimum width=4cm, minimum height=2.25cm] (od) at (\xu-3.5+0.5,\yv+3) {};
\filldraw[color=red!60, fill=red!5, very thick](\xu-4.5+0.5,\yv+3.5) circle (0.5);
\filldraw[color=blue!60, fill=blue!5, very thick](\xu-2.5+0.5,\yv+3.5) circle (0.45);
\filldraw[color=green!60, fill=green!5, very thick](\xu-3.5+0.5,\yv+2.5) circle (0.5);
\node[black] at (\xu-3.5+0.5, \yv+2-0.35) {\small Original $\mathcal{D}$};

\draw[thick, ->] (od.north) to[out=90, in=-90] node[right, align=center]{} (rwt.south);


\node[rectangle, draw, rounded corners, minimum width=0.5cm, minimum height=1.5cm, fill=mygreen!10] (bd1) at (\xu,\yv+6) {\tiny $\mathbf{x}_0$};
\node[rectangle, draw, rounded corners, minimum width=0.5cm, minimum height=1.5cm, fill=mygreen!10] (bd2) at (\xu+1,\yv+6) {\tiny $\mathbf{x}_1$};

\filldraw[black] (\xu+1.25+0.5,\yv+6) circle (2pt);
\filldraw[black] (\xu+1+0.25+0.5+0.5,\yv+6) circle (2pt);
\filldraw[black] (\xu+1.5+1.25,\yv+6) circle (2pt);
\filldraw[black] (\xu+1.5+1.5+0.25,\yv+6) circle (2pt);
\filldraw[black] (\xu+1.5+1.5+0.25+0.5,\yv+6) circle (2pt);

\node[rectangle, draw, rounded corners, minimum width=0.5cm, minimum height=1.5cm, fill=mygreen!10] (bdt2) at (\xu+5,\yv+6) {\tiny$\mathbf{x}_{\scalebox{0.75}{\textit{T-1}}}$};
\node[rectangle, draw, rounded corners, minimum width=0.5cm, minimum height=1.5cm, fill=mygreen!10] (bdt1) at (\xu+6,\yv+6) {\tiny $\mathbf{x}_T$};
\node[rectangle, draw, dashed, rounded corners, minimum width=7cm, minimum height=2.25cm] (fdif) at (\xu+3,\yv+6) {};
\node[black] at (\xu+6, \yv+6) {\small};
\draw[thick, ->] (bd1.east) to[out=0, in=180] node[right, align=center]{} (bd2.west);
\draw[thick, ->] (bdt2.east) to[out=0, in=180] node[right, align=center]{} (bdt1.west);
\node[black] at (\xu+3, \yv+7.35) {\small Forward diffusion};

\draw[thick, ->] (rwt.east) to[out=0, in=180] node[right, align=center]{} (fdif.west);
\end{tikzpicture}
\end{adjustbox}
\end{subfigure}

\caption{Schematic for \name{}. We train our conditional diffusion model using a reweighted objective function. During testing, we condition on the maximum $y$ in the dataset and use classifier-free guidance to sample $Q$ candidate points.
}
\label{fig:modelarch}
\end{figure*}
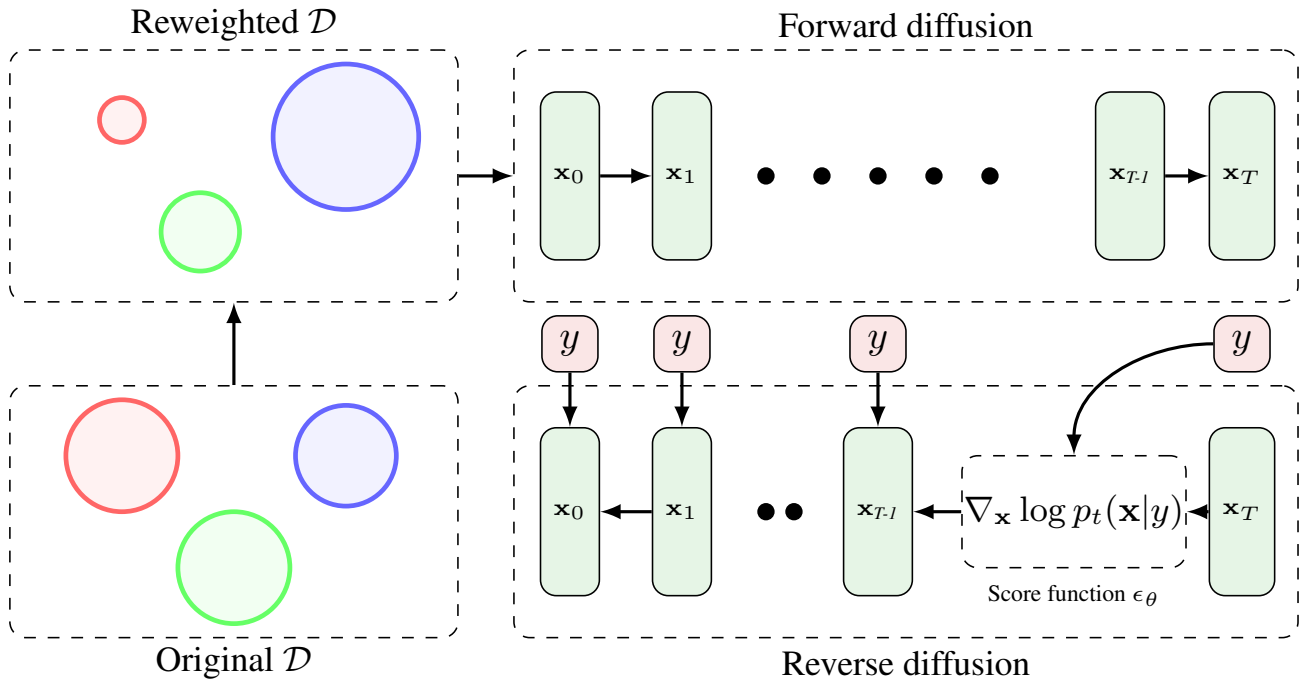

The key challenge for offline BBO is the limited coverage of the offline dataset.
Accordingly, there are two broad classes of approaches in prior works. Following the tradition of online BBO, many \textit{forward} approaches learn a surrogate model mapping inputs to their function values.
Once learned, this surrogate can be optimized with respect to its input using a gradient-based optimizer
or serve as a proxy oracle for a standard online BBO surrogate e.g., BayesOpt using a Gaussian Process.
Forward approaches are indirect, and learning a surrogate that generalizes outside the offline dataset can be challenging~\citep{coms}.

In contrast, \textit{inverse} approaches directly learn a mapping from function values to inputs in the domain of the black-box function.
This mapping is generally one-to-many as many points can have the same function value.
The key advantage of an inverse model is that at test time, we can simply condition the model on high/low function values to obtain candidate optima.
However, learning one-to-many mappings is challenging, especially when the underlying black-box function is defined over a high-dimensional domain.
Prior works have found some success with generative approaches based on generative adversarial networks (GAN)~\citep{mins,GAN,cgan}.
These approaches inherit the problem of their base models, such as training instability and mode collapse for GANs.

We develop a new approach called Denoising Diffusion Optimization Models (\name{}) that uses conditional diffusion models~\citep{ddpm,ddim,scorediff,scorediff2} to parameterize and learn the inverse mapping.
A diffusion model specifies an encoding-decoding procedure based on adding small amounts of noise over multiple timesteps during encoding and then reconstructing the original signal based on the noisy encodings during the decoding stage~\citep{diffusion-original}.
Diffusion models have shown excellent success across a range of continuous data, such as images~\citep{scorediff,ddim,ddpm}, videos~\citep{videodiff}, and speech~\citep{audiodiff1,audiodiff2,diffwave}.
They also allow for conditioning on other inputs, and as a result, they also define flexible models that map one modality to another, such as text2image models~\citep{imagen,dalle}. 
For offline BBO in \name{}, we condition the diffusion model over the observed function values.

 We investigate various design choices that enable successful learning for \name{}. 
In particular, we find there are 2 critical tradeoffs in applying \name{} for offline BBO.
First, even though the offline datapoints are assumed to be sampled i.i.d., we want to specifically condition the model on large values at test-time.
This creates a bias-variance tradeoff where we need to prioritize high value datapoints over others~\citep{mins}.
We resolve this tradeoff by optimizing a weighted loss that downweights the evidence lower bound due to low quality datapoints.

Second, we build on the observation that conditional diffusion models exhibit a diversity-quality tradeoff~\citep{cldiff}.
In emphasizing for diverse candidates, a diffusion model can ignore or downplay the conditioning information.
This is detrimental for offline BBO which explicitly relies on conditioning as a means of optimization at test-time.
Following~\citet{clfreediff}, we fix this issue by decomposing the score function into an unconditional and conditional component. By adjusting the weights of the two components at test-time, we can generate candidates for the optima that explicitly prioritize the conditioning information, as desired.

Empirically, we test \name{} on the Design-Bench suite~\citep{design-bench} of tasks for offline BBO. 
The suite contains a variety of real-world domains spanning both discrete and continuous domains and a variety of data sizes and dimensionalities.
We find that \name{} is very stable to train and performs exceedingly well on this suite. Specifically, \name{} outperforms all the forward and inverse baselines achieving the highest rank on average (\textbf{2.8}).

\section{Background}
\subsection{Problem Statement}
Let $f: \cX \rightarrow \RR$ denote the unknown black-box function, where the domain $\cX$ is an arbitrary subset of $\RR^d$. The black-box optimization problem involves finding a point $\xb^*$ that maximizes $f$:

\begin{equation*}
    \xb^* \in \argmax_{\xb \in \cX} f(\xb)
\end{equation*}

We operate in the setting of \textit{offline} BBO~\citep{design-bench}, where we are not allowed to evaluate $f(\xb)$ for any $\xb$ during training, but must instead make use of an offline dataset of points, which we denote by $\cD  = \{(\xb_1, y_1), (\xb_2, y_2), \cdots, (\xb_n, y_n)\}$. We are allowed a small budget of $Q$ queries during the evaluation to output candidates for the optimal point. 

\subsection{Diffusion Models}
Diffusion models are a class of latent-variable deep generative models that parameterize the encoding and decoding processes via a diffusion process.
The core idea behind diffusion is to add small amounts of noise iteratively to a sample, and train a neural network to invert the transformation. In our work, we make use of continuous time diffusion models~\citep{scorediff,scorediff2}.

Let $\xb_t$ denote a random variable signifying the state of a data point $\xb_0$ at time $t$ where $t\in [0, T]$.
We assume $\xb_0$ is sampled from some unknown data distribution $p_0(\xb)$ and $\xb_T$ represent points sampled from some prior noise distribution (e.g., standard normal distribution).
The forward diffusion process can then be defined as a stochastic differential equation (SDE):
\begin{align}
    d\xb &= \fb(\xb, t)dt + g(t) d\wb
\end{align}
where $\wb$ is the standard Wiener process, $\fb: \RR^d \rightarrow \RR^d$ is the drift coefficient, and $g(t): \RR \rightarrow \RR$ is the diffusion coefficient of $\xb_t$.

The reverse time process (a.k.a. denoising) maps noise to data and can also be defined using an SDE:
\begin{align}
    d\xb &= [\fb(\xb, t) - g(t)^2\nabla_\xb \log p_t(\xb)] dt + g(t) d\tilde{\wb}
\end{align}
where $dt$ is an infinitesimal step backwards in time, and $d\tilde{\wb}$ is a reverse time Wiener process.
In practice, the score function $\nabla_\xb \log p_t(\xb)$ is estimated by a time-dependent neural network $\epsilon_\theta(\xb_t, t)$ trained using a score matching objective such as denoising score matching~\citep{dsm}.


\subsection{Classifier-free Guidance}
In this work, we are interested in training a conditional diffusion model.
In practice, naively conditioning a standard diffusion model by appending the conditioned variable at each step of the denoising process does not work well as the model often ignores or downplays the conditioning information during practice.
One mitigation strategy is based on classifier-free diffusion~\citep{clfreediff}, where we decompose the score function into a linear combination of a conditional and an unconditional score function:
\begin{align}
    \epsilon_\theta(\xb, t, y) = (1+\gamma)\epsilon_{\texttt{cond}}(\xb, t, y) - \gamma\epsilon_{\texttt{uncond}}(\xb, t)
\label{eq:guidance}
\end{align}
Here, the mixing parameter $\gamma$ acts as a trade-off between coverage and fidelity while sampling.
For $\gamma=-1$, the score function translates to sampling from an unconditional diffusion model. For $\gamma\geq0$, the score function prioritizes samples that strongly respect the conditioning information.
Classifier-free guidance obviates the need for training a model on the noisy diffusion data. In practice, instead of learning two separate models for the unconditional and conditional score function, we can train a single model to estimate both by randomly setting the conditioning value to zero during training. 
\section{Denoising Diffusion Optimization Models}
\label{sec:method}
We are interested in learning an inverse model ($y$ to $\xb$ mapping) for offline BBO.
Since $\xb$ is typically high-dimensional, the learned mapping is one-to-many.
We can parameterize any one-to-many mapping as a conditional probability distribution $p(\xb \vert y)$ such that conditioned on a specific value of $y$, the support of the distribution is concentrated on all datapoints $\xb$ for which $f(\xb)\sim y$.

In recent years, deep generative models have shown to excel at learning high-dimensional probability distributions.
For offline BBO, any generative model could be used in principle and related works have explored approaches inspired via generative adversarial networks~\citep{GAN,mins}. 
However, similar to GANs, these BBO approaches are generally hard to train and can also suffer from mode collapse, wherein the sampled points are all very similar.
In this work, we propose to use Diffusion Models for learning the inverse mapping~\citep{diffusion-original}.
Diffusion models have surpassed GANs for many domains, such as image synthesis~\citep{cldiff}. Moreover, they have novel controls for conditional generation, such as guidance, which explicitly allows for trading sample diversity for conditioning.
We refer to our proposed models as \textit{Denoising Diffusion Optimization Models} (\name{}). 

\textbf{Training via Loss Reweighting}
Formally, we train a conditional diffusion model on the offline dataset $\cD$.
During the forward diffusion, we use an SDE with a Variance Preserving (VP) noise perturbation~\citep{scorediff}. Concretely, our forward SDE looks like the following:
\begin{equation}
    d\xb = -\frac{1}{2}\beta_t dt + \sqrt{\beta_t}d\wb
\end{equation}
where $\beta_t = \beta_\texttt{min} + (\beta_\texttt{max} - \beta_\texttt{min})t$ and $t\in [0,1]$. We instantiate the surrogate model for the score function using a simple feed-forward neural network conditioned on the time $t$ and value $y$.
It has been shown previously that a discretization of the above SDE corresponds to the forward diffusion in DDPM~\citep{scorediff,ddpm}.

A naive optimization of the denoising objective would sample $(\xb, y)$ pairs uniformly from the offline dataset. However, we note that since our end goal is to find the argument maximizing $y$, it is important for our model to perform well on relatively high values of $y$.
Filtering out suboptimal $y$ is data inefficient as there could be a learning signal present even when $y$ is low. Instead, we pursue a  reweighting strategy similar to~\citet{mins}. We partition the offline dataset $\cD$ into $N_B$ bins of equal width over $y$. Then for each bin, we assign a weight proportional to the number of points in the bin and the average value of the points in the bin. This ensures that we assign a higher weight during training to (i) bins with more points and (ii) bins with {better} points (points with higher $y$ on average). Concretely the weight $w_i$ for bin $i$ can be computed as:
\begin{equation}
    w_i = \frac{|B_i|}{|B_i| + K} \exp{\left ( \frac{-|\hat{y} - y_{b_i}|}{\tau} \right )}
    \label{eq:reweighing}
\end{equation}
where $\hat{y}$
is the best function value in the offline dataset $\cD$, $|B_i|$ refers to the number of points in the $i^{th}$ bin, and $y_{b_i}$ is the midpoint of the interval corresponding to the bin $B_i$. 
The parameters $K$ and $\tau$ are hyper-parameters, and more details on them can be found in Appendix \ref{app:ablations}.

We finally optimize the objective
\begin{equation}
    \resizebox{0.91\hsize}{!}{$\E[t]{\lambda(t)\E[\xb_0, y]{w(y)\E[\xb_t|\xb_0]{\Vert\epsilon_\theta(\xb_t, t, y) - \nabla_\xb\log p_t(\xb_t | \xb_0)\Vert_2^2}}}$}
    \label{eq:final_obj}
\end{equation}
where $w(y) = w_i$ if $y \in B_i$. Notice the extra reweighting terms introduced here are dependent on $y$ (but independent of $t$ and complementary to the original time-dependent weighting $\lambda(t)$.

During training, we normalize the dimensions of the data points and function values to fit a standard normal. For discrete tasks, we can train a VAE to project the inputs to a continuous domain. However, we map the inputs to log probabilities, following the same procedure as~\citet{design-bench}. They emulate logit values by interpolating between a uniform distribution and the one hot values (this is equivalent to mapping the one hot values to some other constant non-zero values). We find that this simple procedure performs reasonably well in our experiments.

\textbf{Testing via Classifier-Free Guidance}
Once we train the conditional score function, we can use it to generate samples from the reverse SDE. 
Sampling requires a few design decisions: choosing a conditioning value of $y_{\texttt{test}}$, an estimate for the score function, and an SDE solver. Ideally, we would like to choose a $y$ that corresponds to the optima, i.e., $y^\star=f(\xb^\star)$. However, in practice, we do not know $y^\star$; hence, we instead propose to set $y_{\texttt{test}}$ to the maximum value of $y$ in the observed dataset $\cD$, and by setting the guidance weight to a large value. By doing so, we find that the model can generate points further from the unconditional data distribution. This phenomena is analogous to the use of guidance for image generation where sample fidelity can sometimes exceed even real images (e.g., for art) at the cost of diversity. We also find that in practice, this strategy works well, especially with an adjusted score function estimate based on classifier-free guidance (Equation \ref{eq:guidance}).
Finally, we use a second-order Heun solver for sampling from the reverse SDE and find it to work slightly better than first-order solvers in line with recent works~\citep{jolicoeur2021gotta,karras2022elucidating}. 


\begin{algorithm}
\caption{\textbf{D}enoising \textbf{D}iffusion \textbf{O}ptimization \textbf{M}odels
}
\label{alg:bonet}
\textbf{Input} Offline dataset $\cD$, Query budget $Q$, Smoothing parameter $K$, Temperature $\tau$, Number of bins $N_B$ \\
\textbf{Output} A set of proposed candidate points $\mathbf{X}$ with the constraint $|\mathbf{X}| \leq Q$
\begin{algorithmic}[1]
\STATE \COMMENT{\textbf{Phase $1$}: Training}
\STATE Construct bins $\{B_1, \cdots, B_{N_B}\}$ from $\cD$, each bin covering equal $y$-range
\STATE Calculate the weights $(w_1, w_2, \cdots, w_{N_B})$ for each bin using Equation \ref{eq:reweighing} 
\STATE Initialize the model parameters $\theta$
\STATE Train score estimator $\epsilon_\theta$ using Equation \ref{eq:final_obj}
\STATE \COMMENT {\textbf{Phase $2$}: Evaluation }
\STATE $y_{\text{test}} \gets \mathrm{max} \{y \mid (x, y) \in \cD\}$
\STATE $\Xb \gets\phi$
\FOR{$i = 1, \cdots, Q$}
\STATE Sample $\xb_T \sim \cN(0, \Ib)$
\FOR{$t = T-1, \cdots, 0$}
\STATE $\xb_{t} \gets \textsc{HEUN-SAMPLER}(\xb_{t+1}, \theta, y_{\text{test}})$
\ENDFOR
\STATE $\Xb \gets \Xb \cup \{x_0\}$
\ENDFOR\\
\STATE \textbf{return} $\mathbf{X}$

\end{algorithmic}
\end{algorithm}
\section{Experiments}
\subsection{Toy Branin Task}

\begin{wrapfigure}{l}{0.25\textwidth}
\vspace{-2em}
\begin{center}
\includegraphics[scale=0.3]{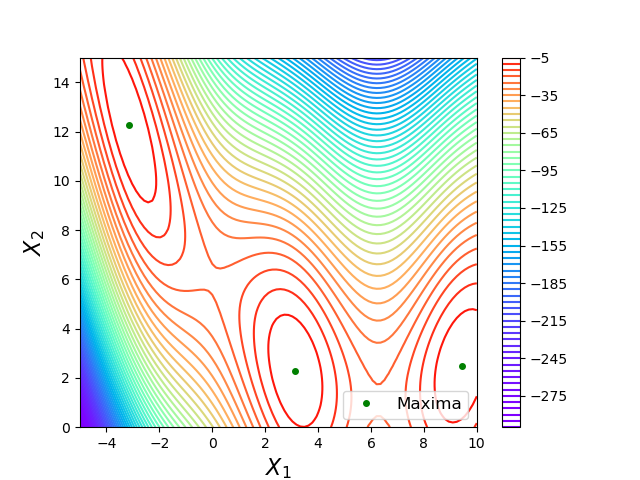}
\end{center}
    \caption{Branin function}
    \label{fig:branin_plot}
\end{wrapfigure}

Branin is a well-known function for benchmarking optimization methods. 
We consider the negative of the standard 2D Branin function in the range $x_1 \in [-5, 10]$ and $x_2 \in [0, 15]$:
\begin{equation}
    f_{br}(x_1, x_2) = -a(x_2 - bx_1^2 + cx_1 - r)^2 - s(1 - t)\cos{x_1} - s
\end{equation}
where $a = 1$, $b = \frac{5.1}{4\pi^2}$, $c = \frac{5}{\pi}$, $r = 6$, $s = 10$, and $t = \frac{1}{8\pi}$. 
In this square region, $f_{br}$ has three global maximas, $(-\pi, 12.275)$, $(\pi, 2.275)$, and $(9.42478, 2.475)$; with the maximum value of $-0.397887$ (Figure \ref{fig:branin_plot}). 
\textbf{Unconditional Diffusion} We first conduct illustrative experiments in an unconditional setting. We emulate an offline dataset that predominantly contains points with high function values and study if \name{} can generate samples from the base distribution. 

\begin{figure*}[!ht]
\centering
\begin{subfigure}{0.12\textwidth}
   \includegraphics[width=\linewidth]{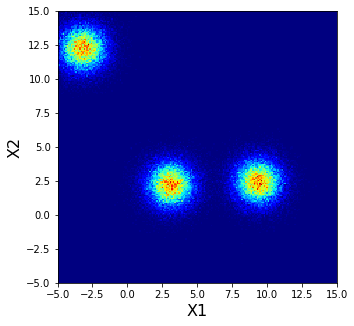} 
   \caption{$\cD_\text{GMM}$}
   \label{fig:data_gmm}
\end{subfigure}
\begin{subfigure}{0.6\textwidth}
   \includegraphics[width=\linewidth]{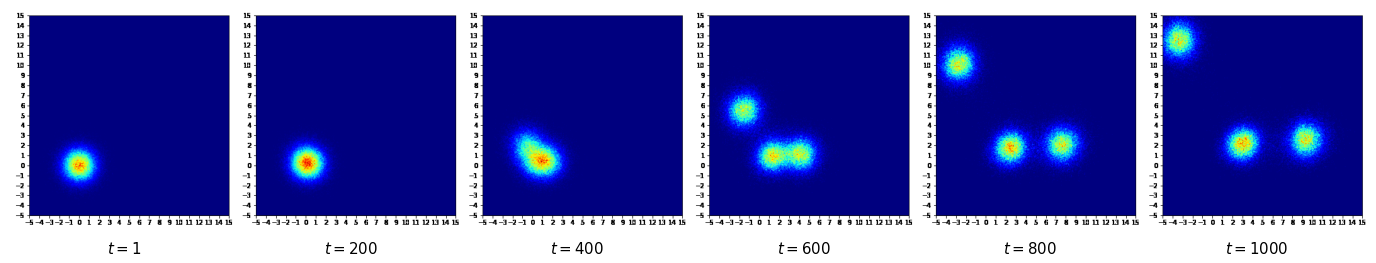} 
   \caption{Sample reverse diffusion trajectory for \name{} trained on $\cD_\text{GMM}$
   }
   \label{fig:traj_gmm}
\end{subfigure}
\begin{subfigure}{0.17\textwidth}
   \includegraphics[width=\linewidth]{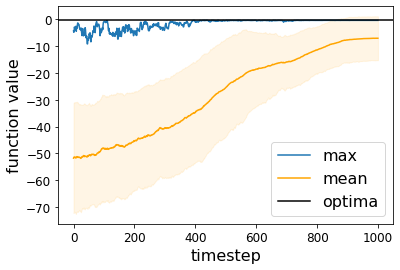}
   \caption{Sample statistics}
   \label{fig:preds_gmm}
\end{subfigure}
\begin{subfigure}{0.12\textwidth}
   \includegraphics[width=\linewidth]{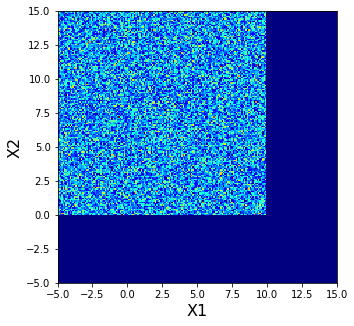}
   \caption{$\cD_\text{Unif}$}
   \label{fig:data_unif}
\end{subfigure}
\begin{subfigure}{0.6\textwidth}
   \includegraphics[width=\linewidth]{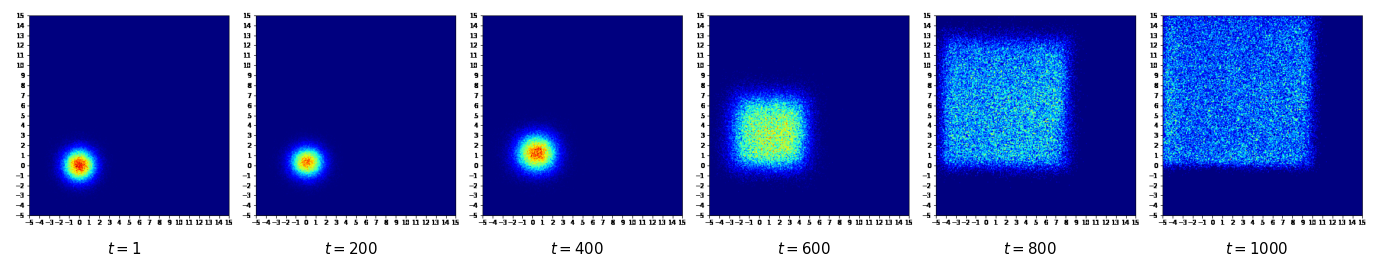} 
   \caption{Sample reverse diffusion trajectory for \name{} trained on $\cD_\text{Unif}$}
   \label{fig:traj_unif}
\end{subfigure}
\begin{subfigure}{0.17\textwidth}
   \includegraphics[width=\linewidth]{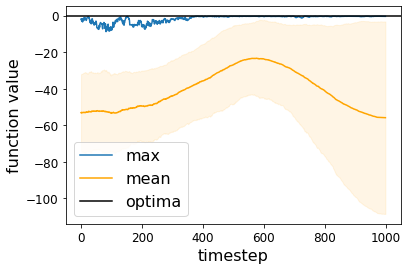} 
   \caption{Sample statistics 
   }
    \label{fig:preds_unif}
\end{subfigure}
\caption{Unconditional diffusion model trained on two data distributions. \textbf{Top row:} (a) Dataset distribution of $\cD_\text{GMM}$. (b) Denoising steps of the diffusion model. (c) Maximum (blue line) and mean (orange line) function value statistics for $256$ points sampled at each timestep in the denoising process, along with the global maxima of Branin (black line). The \textbf{bottom row} shows similar figures for the dataset sampled from a uniform distribution. \name{} is able to reproduce the dataset distribution in both cases. For the GMM case, the mean of the sample set increases, and the variance decreases with increasing timesteps. For the uniform case, the variance increases with increasing timestep, as expected. 
}
\end{figure*}

\begin{figure*}[!ht]
    \centering
    \begin{subfigure}[b]{0.24\textwidth}
    \includegraphics[width=\linewidth]{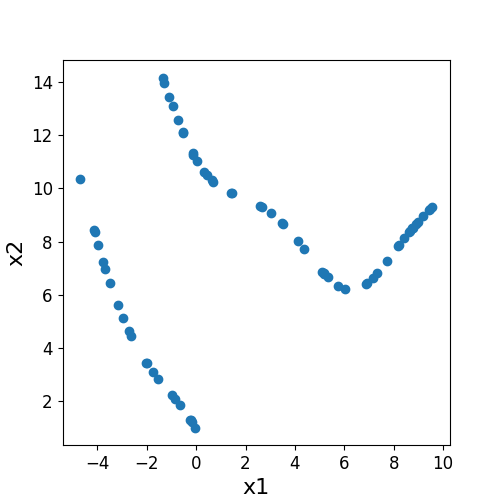}
    \end{subfigure}
    \begin{subfigure}[b]{0.24\textwidth}
    \includegraphics[width=\linewidth]{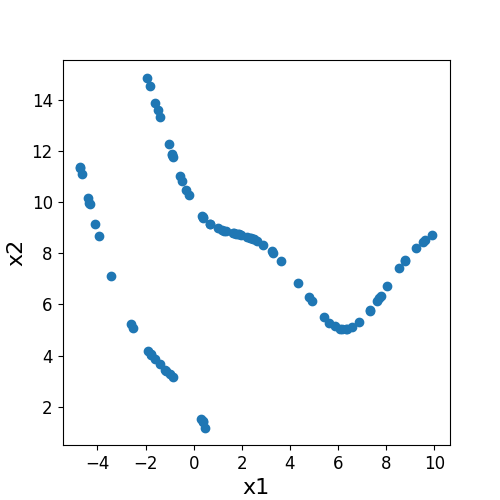}
    \end{subfigure}
    \begin{subfigure}[b]{0.24\textwidth}
    \includegraphics[width=\linewidth]{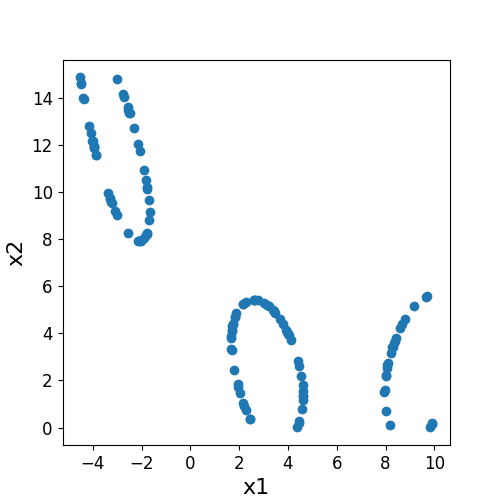}
    \end{subfigure}
    \begin{subfigure}[b]{0.24\textwidth}
    \includegraphics[width=\linewidth]{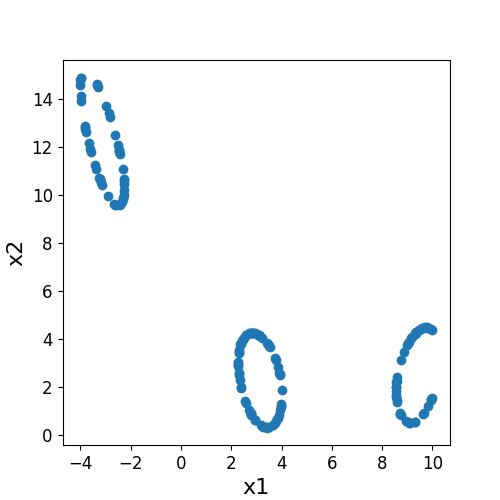}
    \end{subfigure}
     \begin{subfigure}[b]{0.24\textwidth}
    \includegraphics[width=\linewidth]{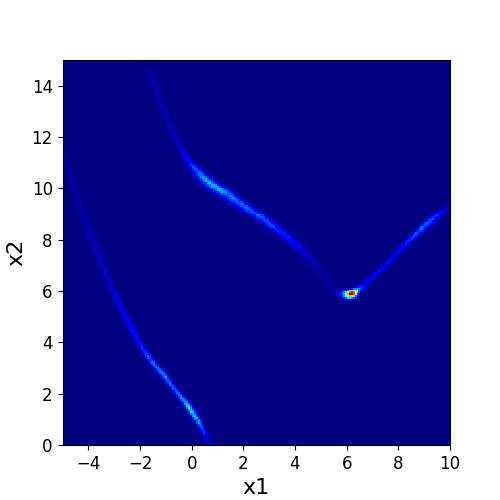}
    \caption{Conditioning $y = 0.2$}
    \end{subfigure}
    \begin{subfigure}[b]{0.24\textwidth}
    \includegraphics[width=\linewidth]{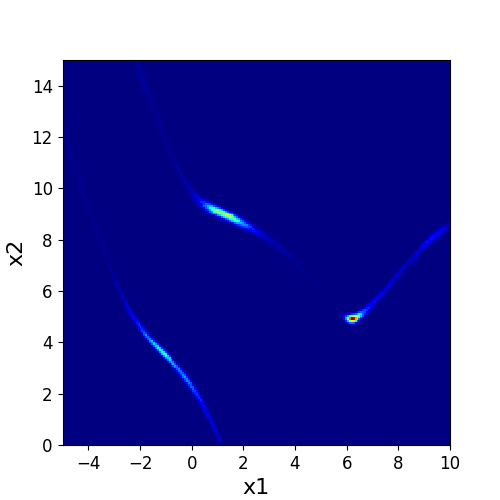}
    \caption{Conditioning $y = 0.4$}
    \end{subfigure}
    \begin{subfigure}[b]{0.24\textwidth}
    \includegraphics[width=\linewidth]{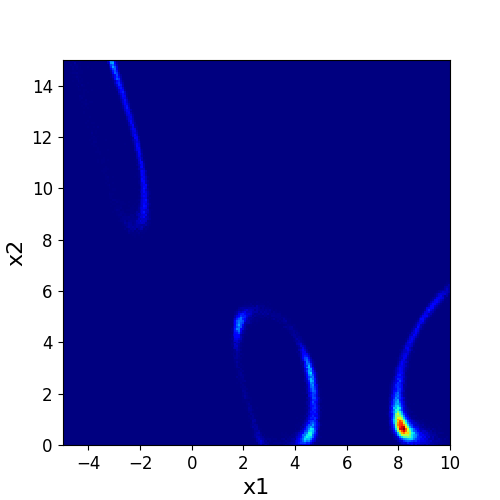}
    \caption{Conditioning $y = 0.9$}
    \end{subfigure}
    \begin{subfigure}[b]{0.24\textwidth}
    \includegraphics[width=\linewidth]{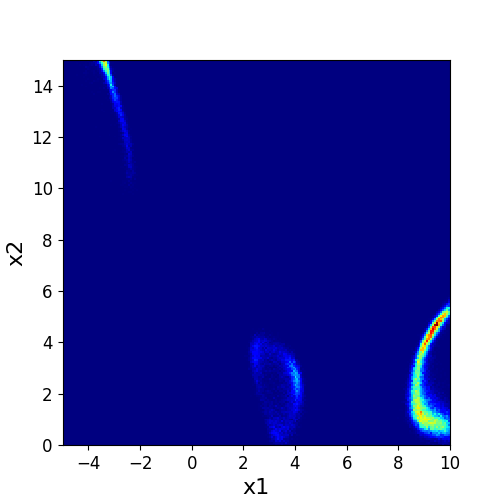}
    \caption{Conditioning $y = 1.0$}

    \end{subfigure}
    \caption{
    Conditional diffusion model trained on $\cD_\text{Unif}$. \textbf{Top row:} ground truth inverse contours found using grid search. \textbf{Bottom row:} the distribution learnt by the conditional diffusion model. Notice the similarity between both plots. Note that the conditioning $y$ value is normalized using the mean and standard deviation of the dataset. 
    }
    \label{fig:inverses}
\end{figure*}

\begin{figure*}[!ht]
\centering
    \begin{subfigure}{0.45\textwidth}
    \centering
    \includegraphics[width=0.9\linewidth]{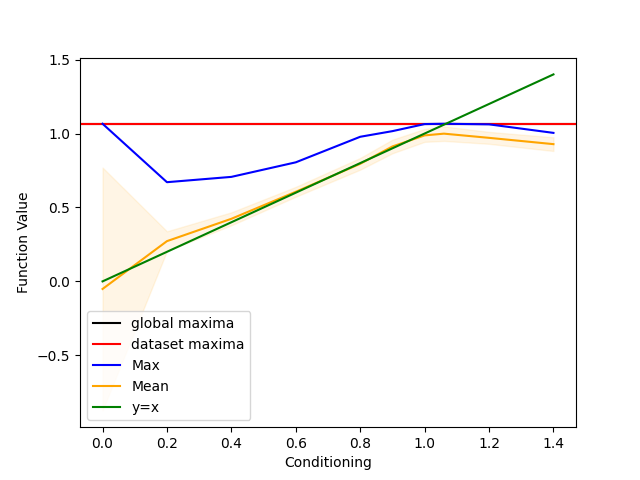} 
   \caption{$\cD_\text{Unif}$}
   \label{fig:preds_unif_cond}
\end{subfigure}
\begin{subfigure}{0.45\textwidth}
\centering\includegraphics[width=0.9\linewidth]{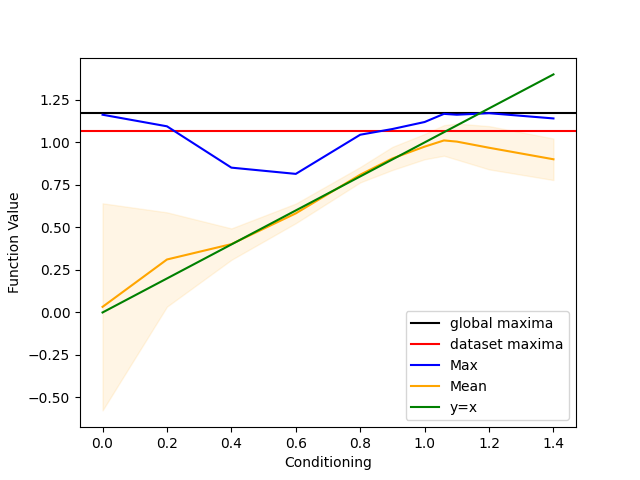} 
   \caption{$\cD_\text{Unif}$ with top $10 \%$-tile points removed}
   \label{fig:preds_unif_trunc_cond}
\end{subfigure}
\caption{Plots of maximum and mean function value of the sampled $256$ points for different normalized conditioning $y$ values for two datasets. Notice that the mean line is very close to the $y = x$ line, as desired. Both the orange and blue line reach their peak when the test-time conditioning is approximately equal to the dataset maxima. }
\end{figure*}
In particular, we sample a dataset $\cD_\text{GMM}$ of size $5000$ points from a 3 component, equi-weighted Gaussian Mixture Model. The means are set as the three global maxima of $f_{br}$ and the covariance matrix for each component is identity (Figure \ref{fig:data_gmm}). We train \name{} with zero conditioning everywhere.
As shown in Figure \ref{fig:traj_gmm},  \name{} can indeed approximate the GMM distribution. Consequently, with increasing timesteps, the denoised samples produced by the model get closer to the optima (Figure \ref{fig:preds_gmm}). 

While the above experiment suggests that diffusion models can indeed be used for offline optimization, it makes an unrealistic assumption on the data distribution being concentrated around high function values by default.
If we consider an alternate dataset $\cD_\text{Unif}$  that is sampled uniformly from the domain (Figure \ref{fig:data_unif}), \name{} learns to reproduce this uniform distribution (Figure \ref{fig:traj_unif}). In this case, the samples generated from the model become increasingly diverse with time, as evident from Figure \ref{fig:preds_unif}. 

\textbf{Conditional Diffusion Model} 
With no control over the dataset quality, we need conditioning mechanisms, as described in Section~\ref{sec:method}.
Consider the dataset $\cD_\text{Unif}$ described above. We now train \name{} conditioned on $y$-values. In Figure \ref{fig:inverses}, we can visualize the contours of the  $f_{br}^{-1}(y)$ learned by our model for different conditioning values $y$. The top row shows the ground truth inverse contours calculated using a simple grid search. The bottom row is the histogram of $100,000$ samples from the last timestep of \name{}. We observe that the learned contours closely match the target contours for all conditioning values of $y$, suggesting an excellent learning of the one-to-many map $f_{br}^{-1}(y)$ in \name{}.
We evaluate the performance of this model in terms of the mean and maximum function values of $256$ samples drawn from the final timestep of the diffusion model. Figure \ref{fig:preds_unif_cond} shows that there is an expected increasing trend up to the conditioning equal to the dataset maxima, and then there is a slight dip. Furthermore, notice that the mean function value vs. conditioning curve is quite close to the $y = x$ line, showing that the model has learned the inverse well.

\textbf{Generalization with Suboptimal Datasets} The datasets we considered so far contain points close to the maxima. A key motivation for offline BBO is to generalize beyond the given dataset. To test this property for \name{}, we create a more challenging task by removing the top $10-\%$tile points from $\cD_\text{Unif}$ based on their function values. We again train a conditional diffusion model and plot our predictions in Figure \ref{fig:preds_unif_trunc_cond}. We observe that the model is able to successfully propose points with function values higher than the dataset maximum. Further, the peak performance (in terms of both the best and mean function values) is when the conditioning is the maximum function value in the dataset.

    
\subsection{Design-Bench}
We also test \name{} on 6 high-dimensional real-world tasks in Design-Bench~\citep{design-bench}, a suite of offline BBO tasks. We test on three continuous and three discrete tasks\footnote{We exclude NAS as it requires excessive compute  beyond our resources for evaluating across multiple seeds. We exclude Hopper as this domain is known to be buggy. See details in Appendix \ref{app::impact}.}.  In \textbf{D'Kitty} and \textbf{Ant} Morphology, we need to optimize for the morphology of robots.
 In \textbf{Superconductor (Supercond.)}, the aim is to optimize for finding a superconducting material with a high critical temperature.
\textbf{TFBind8} and \textbf{TFBind10} are discrete tasks where the goal is to find a DNA sequence that has a maximum affinity to bind with a specified transcription factor.
 \textbf{ChEMBL} is a discrete task that optimizes drugs for specific chemical properties.
\begin{table*}[ht!]
    \centering
    \resizebox{\textwidth}{!}{
\begin{tabular}{llllllllc}
\toprule
\multicolumn{1}{c}{\textbf{BASELINE}}  &\multicolumn{1}{c}{\textbf{TFBIND8}} &\multicolumn{1}{c}{\textbf{TFBIND10}} &\multicolumn{1}{c}{\textbf{SUPERCON.}} &\multicolumn{1}{c}{\textbf{ANT}}  &\multicolumn{1}{c}{\textbf{D'KITTY}} &\multicolumn{1}{c}{\textbf{CHEMBL}}  & \multicolumn{1}{c}{\textbf{MEAN SCORE}} & \multicolumn{1}{c}{\textbf{MEAN RANK}}\\
\midrule
$\cD$ (best) & $0.439$ & $0.467$ & $0.399$ & $0.565$ & $0.884$ & $0.605$ & {-} & {-}\\
\midrule
CbAS & $0.958 \pm 0.018$  & $0.657 \pm 0.017$  & $0.45 \pm 0.083$  & $0.876 \pm 0.015$  & $0.896 \pm 0.016$ & $0.640 \pm 0.005$ & $0.746 \pm 0.003$ & $5.5$ \\ 
GP-qEI & $0.824 \pm 0.086$  & $0.635 \pm 0.011$  & $0.501 \pm 0.021$  & $0.887 \pm 0.0$  & $0.896 \pm 0.0$ & $0.633 \pm 0.000$ & $0.729 \pm 0.019$ & $6.2$ \\ 
CMA-ES & $0.933 \pm 0.035$  & $\boldmathred{0.679 \pm 0.034}$  & $0.491 \pm 0.004$  & $\boldmathblue{1.436 \pm 0.928}$  & $0.725 \pm 0.002$ & $0.636 \pm 0.004$ & $\boldmathblue{0.816 \pm 0.168}$ & $4.5$ \\ 
Gradient Ascent & $\boldmathblue{0.981 \pm 0.015}$  & $0.659 \pm 0.039$  & $\boldmathred{0.504 \pm 0.005}$  & $0.34 \pm 0.034$  & $0.906 \pm 0.017$ & $0.647 \pm 0.020$ & $0.672 \pm 0.021$ & $\boldmathred{3.5}$ \\ 
REINFORCE & $0.959 \pm 0.013$  & $0.64 \pm 0.028$  & $0.481 \pm 0.017$  & $0.261 \pm 0.042$  & $0.474 \pm 0.202$ & $0.636 \pm 0.023$ & $0.575 \pm 0.054$ & $6.3$ \\ 
MINs & $0.938 \pm 0.047$  & $0.659 \pm 0.044$  & $0.484 \pm 0.017$  & $0.942 \pm 0.018$  & $\boldmathred{0.944 \pm 0.009}$ & $\boldmathblue{0.653 \pm 0.002}$ & $0.770\pm 0.023$ & $\boldmathred{3.5}$ \\
COMs & $0.964 \pm 0.02$  & $0.654 \pm 0.02$  & $0.423 \pm 0.033$  & $0.949 \pm 0.021$  & $\boldmathblue{0.948 \pm 0.006}$ & $\boldmathred{0.648 \pm 0.005}$ & $0.764\pm 0.018$ & $3.7$ \\
\midrule
\name{} & $\boldmathred{0.971 \pm 0.005}$ & $\boldmathblue{0.688 \pm 0.092}$ & $\boldmathblue{0.560 \pm 0.044}$ & $\boldmathred{0.957 \pm 0.012}$ & $0.926 \pm 0.009$ & $0.633 \pm 0.007$ & $\boldmathred{0.787 \pm 0.034}$ & $\boldmathblue{2.8}$\\
\bottomrule
\end{tabular}}
    \caption{Comparative evaluation of \name{} over 6 tasks, with each task averaged over 5 seeds. We report normalized results (along with stddev) with a budget $Q = 256$. We highlight the top two results in each column (and where there is a tie, we highlight both). $\boldmathblue{Blue}$ refers to the best entry, and $\boldmathred{Violet}$ refers to the second best. We find that \name{} achieves the best average rank of all the baselines and is second best on the mean score metric.}
    \label{tab:main_table}
\end{table*}

\begin{table*}[h]
    \centering
    \resizebox{\textwidth}{!}{
\begin{tabular}{lllllll}
\toprule
{} &             \textbf{D'KITTY} &                \textbf{ANT} &          \textbf{TFBIND8} &         \textbf{TFBIND10} &     \textbf{SUPERCON.} &             \textbf{CHEMBL} \\
\midrule
No reweighting &  $0.926 \pm 0.008$ &  $0.888 \pm 0.018$ & $0.957 \pm 0.000$ &  $0.644 \pm 0.011$ &  $0.553 \pm 0.051$ & $\mathbf{0.633 \pm 0.000}$\\
Reweighting                                  &  $\mathbf{0.930 \pm 0.003}$ &  $\mathbf{0.960 \pm 0.015}$ &  $\mathbf{0.971 \pm 0.005}$ &  $\mathbf{0.688 \pm 0.092}$ &  $\mathbf{0.560 \pm 0.044}$ &  $\mathbf{0.633 \pm 0.000}$ \\
\bottomrule
\end{tabular}}

    \caption{Comparison of normalized scores for \name{} with and without reweighting on Design-Bench. Higher is better We find that there is a significant improvement in score from reweighting on most tasks. We report scores averaged across 5 seeds.}
    \label{tab:reweighting}
\end{table*}
\textbf{Baselines}
We compare \name{} with multiple baselines using canonical approaches like gradient ascent, Bayesian Optimization (BayesOpt), REINFORCE~\citep{reinforce}, evolutionary strategies like CMA-ES~\citep{hansen2006cma},  and newer methods like MINs~\citep{mins}, COMs~\citep{coms} and CbAS~\citep{cbas}. Since we are in the offline setting, for active methods like BayesOpt, we follow the procedure of \citet{design-bench} and perform BayesOpt on a surrogate model $\hat{f}(\xb)$ trained on the offline dataset. We instantiate BayesOpt using a Gaussian Process as the uncertainty quantifier and the quasi-Expected Improvement (q-EI) acquisition function. For all tasks, we use a query budget $Q = 256$. Following the procedure used by \citet{design-bench}, we report the results of all the tasks normalized using a larger unseen offline dataset, i.e. we report $y_{\texttt{norm}}$ where $y_{\texttt{norm}} = \frac{y - y_{\texttt{min}}}{y_{\texttt{max}}-y_{\texttt{min}}}$, where $y$, $y_{\texttt{min}}$ and $y_{\texttt{max}}$ refer to the score of the proposed solution, and the minimum and maximum of the large unseen offline dataset. Note that this larger dataset is \textit{not} used for training but only for reporting normalized results. We also report the mean score and mean rank of all baselines.

\textbf{Architecture} We instantiate \name{} using a simple feed-forward neural network with 2 hidden layers, width of $1024$ and ReLU activation. We train using a fixed learning rate of $0.001$ and batch size of $128$. We set the minimum and maximum noise variance to be $0.01$ and $2.0$ respectively. We use the same value of $\gamma = 2.0$ across all experiments. 

\textbf{Main Results}
In Table \ref{tab:main_table}, we report normalized results of the max score achieved by \name{} and the baselines along with the mean normalized score and the mean rank. Overall we find that \name{} achieves an average rank of $2.8$, the best among all the baselines, and an average score of $0.787$. We achieve the best result on 2 tasks and are runner-up on another 2 tasks. On Superconductor, we outperform other baselines by a significant margin, beating the next closest baseline by 11\%. We further note that CMA-ES, the baseline with the highest average score, has a standard deviation $5$ times as large as \name{}, indicating that it is very sensitive to initialization, unlike \name{}.

\subsection{Ablations}
\label{sec:conditioning}
We perform ablation studies on the different components of \name{} to study their effects: reweighting, conditioning, and classifier-free diffusion. We provide additional results and discussion in Appendix \ref{app:ablations}.

\begin{figure}[!ht]
\centering
\begin{subfigure}{0.5\textwidth}
   \includegraphics[width=0.9\linewidth]{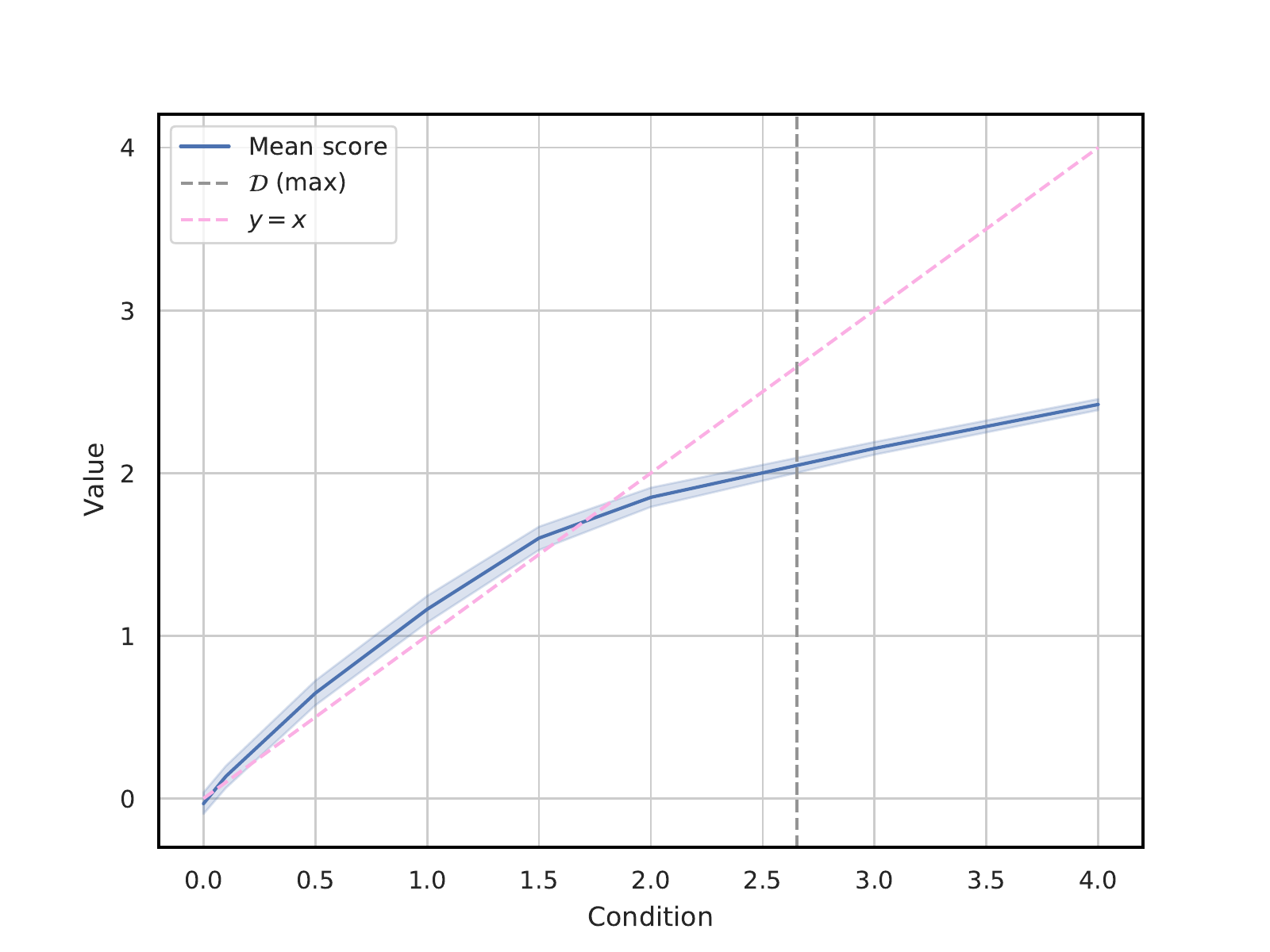} 
\end{subfigure}
\caption{Plot of mean function value versus conditioning for Superconductor. Dotted lines represent the dataset max and the line $y = x$ (the ideal line). Even for such a high-dimensional task, a strong correlation exists between predicted and conditioned values.}
\label{fig:cond}
\end{figure}
\begin{figure}
    \centering
    \begin{subfigure}{0.5\textwidth}
    \includegraphics[width=0.9\linewidth]{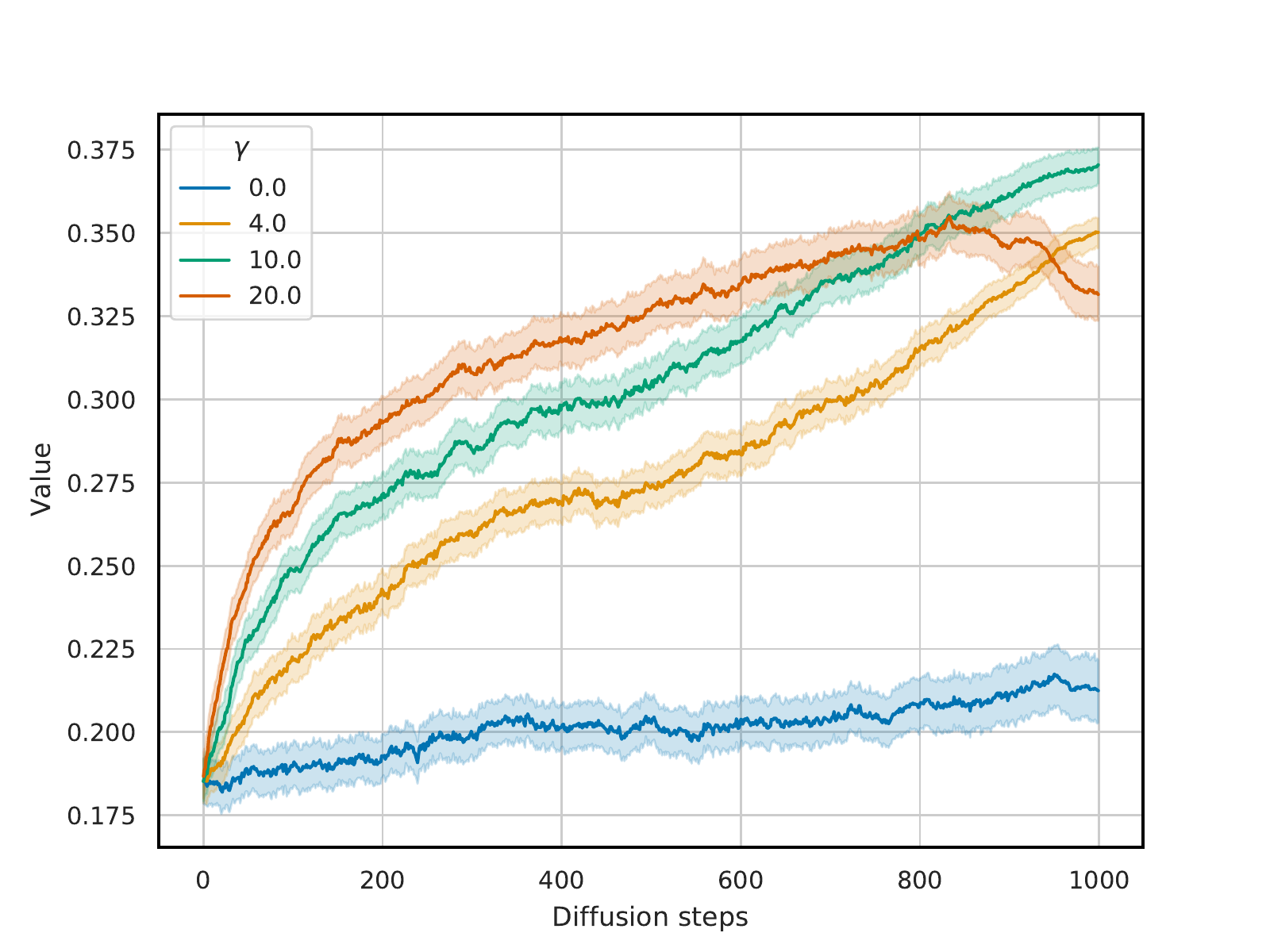}
    \end{subfigure}
    \caption{Change in mean objective values as a function of number of diffusion steps for multiple values of $\gamma$ for Superconductor. We find that when there is no guidance ($\gamma = 0$), \name{} performs poorly, indicating the importance of guidance.
    }
    \label{fig:gamma}
\end{figure}
\textbf{Reweighting} We report results with and without reweighting in \name{} in Table \ref{tab:reweighting} across 6 tasks. We find that across all tasks, the reweighted model outperforms the non-reweighted one, indicating the importance of reweighting.

\textbf{Impact of conditioning} In Figure \ref{fig:cond}, we plot the mean of the predicted values against the conditioned value for the Superconductor task. We see that up to around the maxima of the dataset, both the conditioned and predicted values correlate very well with each other, suggesting that increasing the conditioning also increases the score of the predicted point. While generalizing beyond the dataset maxima is difficult when dealing with such a high-dimensional task, we notice a fairly high correlation between predicted and conditioned values even for very large conditioning values. 

\textbf{Impact of guidance} We perform an ablation on the impact of the weighing factor in classifier-free guidance in Figure \ref{fig:gamma}. We plot the average score of 512 samples from our model at various timesteps during the diffusion process for Superconductor. As expected, we find that no guidance $\gamma = 0$ (i.e., a simple conditional diffusion model) performs very poorly compared to models with guidance, indicating the importance of classifier-free guidance. We also notice that increasing the guidance also increases the rate at which the predicted values increase, indicating the importance of using higher guidance weights during optimization.


\section{Related Work}
\textbf{Active BBO} Most prior work in BBO has been in the active setting, where models can query the function during training. Bayesian optimization is the most well-known active BBO method, and there is a large body of prior work in the area~\citep{snoek_bayesopt,nando-bayesopt,ICML-2010-SrinivasKKS,zhang2022unifying} \textit{inter alia}. Such methods usually use surrogates like Gaussian Processes (GPs) to model the underlying function and sequentially update it by querying new points determined by an uncertainty-aware acquisition function. Bandit algorithms~\citep{JMLR:v16:swaminathan15a_bandit,bandit_exp_then_commit,aistats18b,joachims2018deep_bandit,riquelme2018deepbandit,attia2019closed,grover_bandits} are another well-known class of active BBO methods. \name{} is similar to Neural Diffusion Processes (NDP)~\citep{ndp} in the use of diffusion processes for black-box optimization but  operates in a purely offline setting, imposes no restrictions on the model architecture, and uses classifier-free guidance for generalization, unlike NDPs.

\textbf{Offline BBO} Recent works have shown significant progress in the offline setting~\citep{coms, mins, hansen2006cma, autocbas}. ~\citet{design-bench} provides Design-Bench, a comprehensive real-world dataset for benchmarking offline BBO methods. 
Most methods~\citep{coms, design-bench,nguyen2022transformer} use forward models to approximate the black-box function and optimize it. On the other hand, methods like~\citet{mins} train generative models like GANs to model the one-to-many inverse mapping from function value to the input domain. CbAS~\citep{cbas} learns a density model in the space of inputs that approximates the data distribution, and gradually adapts it towards an optimized solution. In particular CbAS alternates between training a VAE on a set of samples and generating new samples from the autoencoder which are then used to train the VAE on the next iteration.

\textbf{Diffusion Models} Diffusion models~\citep{diffusion-original} have recently shown strong performance in robust generative modeling for images~\citep{ddim, scorediff, scorediff2, ddpm}, videos~\citep{videodiff}, and audio~\citep{audiodiff1, audiodiff2}, as well as for planning in RL~\citep{janner2022planning}.
The latter is particularly relevant and includes parallel advancements in the use of autoregressive and masked transformers that reduce sequential decision making to a generative modeling problem~\cite{chen2021decisiontransformer,liu2022masked,zhuscaling,nguyen2022reliable,onlinedt,zheng2022semi}.
Usually a model is trained to invert the forward diffusion process of converting data to noise sequentially by adding small Gaussian noise at each timestep.
Our work can benefit from this growing area of research spanning advancements in conditioning and sampling strategies~\citep{jolicoeur2021gotta,karras2022elucidating,bansal2023leaving} and alternate diffusion parameterizations~\citep{bansal2022cold,hoogeboom2022blurring}, among others.
\section{Summary}
We proposed Denoising Diffusion Optimization Models (\name{}), a new approach for offline black-box optimization based on conditional diffusion models.
\name{} falls within the category of inverse approaches that train a mapping from function values to points.
Unlike prior inverse approaches such as MINs~\citep{mins}, \name{} is more stable to train and less susceptible to mode collapse.
We noted two key modifications to the default setup: the use of loss reweighting to prioritize offline points with high function values and the use of classifier-guided sampling for improving the strength of conditioning at test time.
Empirically, our approach is highly competitive and on average, outperforms both forward and inverse baselines on the Design-Bench suite~\citep{design-bench}.

\textbf{Limitations and Future Work.} 
Diffusion models is that they are relatively slow to sample compared to other generative approaches.
While sampling speed is typically not a major bottleneck for offline BBO, it can be potentially limiting for some real-time applications.
Given the fast-developing field of diffusion models for data generation, we expect many of the corresponding advancements in sampling and hyperparameter tuning~\citep{karras2022elucidating}  to transfer over to \name{}.
Finally, we focused on offline BBO in this work. In some practical scenarios, we might also have resources for active data acquisition and future work  can explore using \name{} to warm start a subsequent diffusion-based online BBO method.

\section*{Acknowledgements}
We are grateful to Tung Nguyen for helpful comments on early drafts of this work. This work used computational and storage services associated with the Hoffman2 Shared Cluster provided by UCLA Institute for Digital Research and Education’s Research Technology Group. Aditya Grover was supported in part by a Cisco Faculty Award and a Sony Faculty Innovation Award.

\bibliography{example_paper}

\begin{thebibliography}{50}
\providecommand{\natexlab}[1]{#1}
\providecommand{\url}[1]{\texttt{#1}}
\expandafter\ifx\csname urlstyle\endcsname\relax
  \providecommand{\doi}[1]{doi: #1}\else
  \providecommand{\doi}{doi: \begingroup \urlstyle{rm}\Url}\fi

\bibitem[Attia et~al.(2020)Attia, Grover, Jin, Severson, Cheong, Liao, Chen,
  Perkins, Yang, Herring, Aykol, Harris, Braatz, Ermon, and
  Chueh]{attia2019closed}
Attia, P., Grover, A., Jin, N., Severson, K., Cheong, B., Liao, J., Chen,
  M.~H., Perkins, N., Yang, Z., Herring, P., Aykol, M., Harris, S., Braatz, R.,
  Ermon, S., and Chueh, W.
\newblock Closed-loop optimization of extreme fast charging for batteries using
  machine learning.
\newblock \emph{Nature}, 2020.

\bibitem[Bansal et~al.(2022)Bansal, Borgnia, Chu, Li, Kazemi, Huang, Goldblum,
  Geiping, and Goldstein]{bansal2022cold}
Bansal, A., Borgnia, E., Chu, H.-M., Li, J.~S., Kazemi, H., Huang, F.,
  Goldblum, M., Geiping, J., and Goldstein, T.
\newblock Cold diffusion: Inverting arbitrary image transforms without noise.
\newblock \emph{arXiv preprint arXiv:2208.09392}, 2022.

\bibitem[Bansal \& Grover(2023)Bansal and Grover]{bansal2023leaving}
Bansal, H. and Grover, A.
\newblock Leaving reality to imagination: Robust classification via generated
  datasets.
\newblock \emph{arXiv preprint arXiv:2302.02503}, 2023.

\bibitem[Brookes et~al.(2019)Brookes, Park, and Listgarten]{cbas}
Brookes, D., Park, H., and Listgarten, J.
\newblock Conditioning by adaptive sampling for robust design.
\newblock In \emph{International Conference on Machine Learning}, 2019.

\bibitem[Chen et~al.(2021)Chen, Lu, Rajeswaran, Lee, Grover, Laskin, Abbeel,
  Srinivas, and Mordatch]{chen2021decisiontransformer}
Chen, L., Lu, K., Rajeswaran, A., Lee, K., Grover, A., Laskin, M., Abbeel, P.,
  Srinivas, A., and Mordatch, I.
\newblock Decision transformer: Reinforcement learning via sequence modeling.
\newblock \emph{arXiv preprint arXiv:2106.01345}, 2021.

\bibitem[Dhariwal \& Nichol(2021)Dhariwal and Nichol]{cldiff}
Dhariwal, P. and Nichol, A.~Q.
\newblock Diffusion models beat {GAN}s on image synthesis.
\newblock In \emph{Advances in Neural Information Processing Systems}, 2021.

\bibitem[Dutordoir et~al.(2022)Dutordoir, Saul, Ghahramani, and Simpson]{ndp}
Dutordoir, V., Saul, A., Ghahramani, Z., and Simpson, F.
\newblock Neural diffusion processes.
\newblock \emph{arXiv preprint arXiv:2206.03992}, 2022.

\bibitem[Fannjiang \& Listgarten(2020)Fannjiang and Listgarten]{autocbas}
Fannjiang, C. and Listgarten, J.
\newblock Autofocused oracles for model-based design.
\newblock In \emph{Advances in Neural Information Processing Systems},
  volume~33, 2020.

\bibitem[Garivier et~al.(2016)Garivier, Kaufmann, and
  Lattimore]{bandit_exp_then_commit}
Garivier, A., Kaufmann, E., and Lattimore, T.
\newblock On explore-then-commit strategies.
\newblock In \emph{Proceedings of the 30th International Conference on Neural
  Information Processing Systems}, 2016.

\bibitem[Goodfellow et~al.(2014)Goodfellow, Pouget-Abadie, Mirza, Xu,
  Warde-Farley, Ozair, Courville, and Bengio]{GAN}
Goodfellow, I., Pouget-Abadie, J., Mirza, M., Xu, B., Warde-Farley, D., Ozair,
  S., Courville, A., and Bengio, Y.
\newblock Generative adversarial nets.
\newblock In \emph{Advances in Neural Information Processing Systems}, 2014.

\bibitem[Grover et~al.(2018)Grover, Markov, Attia, Jin, Perkins, Cheong, Chen,
  Yang, Harris, Chueh, and Ermon]{aistats18b}
Grover, A., Markov, T., Attia, P., Jin, N., Perkins, N., Cheong, B., Chen, M.,
  Yang, Z., Harris, S., Chueh, W., and Ermon, S.
\newblock Best arm identification in multi-armed bandits with delayed feedback.
\newblock In \emph{International Conference on Artificial Intelligence and
  Statistics (AISTATS)}, 2018.

\bibitem[Guo et~al.(2021)Guo, Agrawal, Grover, Muthukumar, and
  Pananjady]{grover_bandits}
Guo, W., Agrawal, K.~K., Grover, A., Muthukumar, V., and Pananjady, A.
\newblock Learning from an exploring demonstrator: Optimal reward estimation
  for bandits.
\newblock In \emph{International Conference on Artificial Intelligence and
  Statistics (AISTATS)}, 2021.

\bibitem[Hansen(2006)]{hansen2006cma}
Hansen, N.
\newblock The cma evolution strategy: a comparing review.
\newblock \emph{Towards a new evolutionary computation: Advances in the
  estimation of distribution algorithms}, pp.\  75--102, 2006.

\bibitem[Hansen(2016)]{hansen2016cma}
Hansen, N.
\newblock The cma evolution strategy: A tutorial.
\newblock \emph{arXiv preprint arXiv:1604.00772}, 2016.

\bibitem[Ho \& Salimans(2021)Ho and Salimans]{clfreediff}
Ho, J. and Salimans, T.
\newblock Classifier-free diffusion guidance.
\newblock In \emph{NeurIPS 2021 Workshop on Deep Generative Models and
  Downstream Applications}, 2021.

\bibitem[Ho et~al.(2020)Ho, Jain, and Abbeel]{ddpm}
Ho, J., Jain, A., and Abbeel, P.
\newblock Denoising diffusion probabilistic models.
\newblock In \emph{Advances in Neural Information Processing Systems}, 2020.

\bibitem[Ho et~al.(2022)Ho, Salimans, Gritsenko, Chan, Norouzi, and
  Fleet]{videodiff}
Ho, J., Salimans, T., Gritsenko, A.~A., Chan, W., Norouzi, M., and Fleet, D.~J.
\newblock Video diffusion models.
\newblock In \emph{Advances in Neural Information Processing Systems}, 2022.

\bibitem[Hoogeboom \& Salimans(2022)Hoogeboom and
  Salimans]{hoogeboom2022blurring}
Hoogeboom, E. and Salimans, T.
\newblock Blurring diffusion models.
\newblock \emph{arXiv preprint arXiv:2209.05557}, 2022.

\bibitem[Huang et~al.(2021)Huang, Lim, and Courville]{scorediff2}
Huang, C.-W., Lim, J.~H., and Courville, A.~C.
\newblock A variational perspective on diffusion-based generative models and
  score matching.
\newblock \emph{Advances in Neural Information Processing Systems},
  34:\penalty0 22863--22876, 2021.

\bibitem[Janner et~al.(2022)Janner, Du, Tenenbaum, and
  Levine]{janner2022planning}
Janner, M., Du, Y., Tenenbaum, J.~B., and Levine, S.
\newblock Planning with diffusion for flexible behavior synthesis.
\newblock \emph{arXiv preprint arXiv:2205.09991}, 2022.

\bibitem[Jeong et~al.(2021)Jeong, Kim, Cheon, Choi, and Kim]{audiodiff2}
Jeong, M., Kim, H., Cheon, S.~J., Choi, B.~J., and Kim, N.~S.
\newblock Diff-tts: A denoising diffusion model for text-to-speech.
\newblock \emph{arXiv preprint arXiv:2104.01409}, 2021.

\bibitem[Joachims et~al.(2018)Joachims, Swaminathan, and
  de~Rijke]{joachims2018deep_bandit}
Joachims, T., Swaminathan, A., and de~Rijke, M.
\newblock Deep learning with logged bandit feedback.
\newblock In \emph{International Conference on Learning Representations}, 2018.

\bibitem[Jolicoeur-Martineau et~al.(2021)Jolicoeur-Martineau, Li,
  Pich{\'e}-Taillefer, Kachman, and Mitliagkas]{jolicoeur2021gotta}
Jolicoeur-Martineau, A., Li, K., Pich{\'e}-Taillefer, R., Kachman, T., and
  Mitliagkas, I.
\newblock Gotta go fast when generating data with score-based models.
\newblock \emph{arXiv preprint arXiv:2105.14080}, 2021.

\bibitem[Karras et~al.(2022)Karras, Aittala, Aila, and
  Laine]{karras2022elucidating}
Karras, T., Aittala, M., Aila, T., and Laine, S.
\newblock Elucidating the design space of diffusion-based generative models.
\newblock \emph{arXiv preprint arXiv:2206.00364}, 2022.

\bibitem[Kim et~al.(2022)Kim, Kim, and Yoon]{audiodiff1}
Kim, H., Kim, S., and Yoon, S.
\newblock Guided-tts: A diffusion model for text-to-speech via classifier
  guidance.
\newblock In \emph{International Conference on Machine Learning}. PMLR, 2022.

\bibitem[Kong et~al.(2021)Kong, Ping, Huang, Zhao, and Catanzaro]{diffwave}
Kong, Z., Ping, W., Huang, J., Zhao, K., and Catanzaro, B.
\newblock Diffwave: A versatile diffusion model for audio synthesis.
\newblock In \emph{International Conference on Learning Representations}, 2021.

\bibitem[Krishnamoorthy et~al.(2023)Krishnamoorthy, Mashkaria, and
  Grover]{bonet}
Krishnamoorthy, S., Mashkaria, S.~M., and Grover, A.
\newblock Generative pretraining for black-box optimization.
\newblock In \emph{International Conference on Machine Learning (ICML)}, 2023.

\bibitem[Kumar \& Levine(2020)Kumar and Levine]{mins}
Kumar, A. and Levine, S.
\newblock Model inversion networks for model-based optimization.
\newblock In \emph{Advances in Neural Information Processing Systems}, 2020.

\bibitem[Liu et~al.(2022)Liu, Liu, Grover, and Abbeel]{liu2022masked}
Liu, F., Liu, H., Grover, A., and Abbeel, P.
\newblock Masked autoencoding for scalable and generalizable decision making.
\newblock In \emph{Advances in Neural Information Processing Systems
  (NeurIPS)}, 2022.

\bibitem[Mirza \& Osindero(2014)Mirza and Osindero]{cgan}
Mirza, M. and Osindero, S.
\newblock Conditional generative adversarial nets.
\newblock \emph{arXiv preprint arXiv:1411.1784}, 2014.

\bibitem[Nguyen \& Grover(2022)Nguyen and Grover]{nguyen2022transformer}
Nguyen, T. and Grover, A.
\newblock Transformer neural processes: Uncertainty-aware meta learning via
  sequence modeling.
\newblock In \emph{International Conference on Machine Learning (ICML)}, 2022.

\bibitem[Nguyen et~al.(2022)Nguyen, Zheng, and Grover]{nguyen2022reliable}
Nguyen, T., Zheng, Q., and Grover, A.
\newblock Reliable conditioning of behavioral cloning for offline reinforcement
  learning.
\newblock \emph{arXiv preprint arXiv:2210.05158}, 2022.

\bibitem[Ramesh et~al.(2021)Ramesh, Pavlov, Goh, Gray, Voss, Radford, Chen, and
  Sutskever]{dalle}
Ramesh, A., Pavlov, M., Goh, G., Gray, S., Voss, C., Radford, A., Chen, M., and
  Sutskever, I.
\newblock Zero-shot text-to-image generation.
\newblock In \emph{International Conference on Machine Learning}, 2021.

\bibitem[Riquelme et~al.(2018)Riquelme, Tucker, and
  Snoek]{riquelme2018deepbandit}
Riquelme, C., Tucker, G., and Snoek, J.
\newblock Deep bayesian bandits showdown: An empirical comparison of bayesian
  deep networks for thompson sampling.
\newblock \emph{arXiv preprint arXiv:1802.09127}, 2018.

\bibitem[Saharia et~al.(2022)Saharia, Chan, Saxena, Li, Whang, Denton,
  Ghasemipour, Gontijo~Lopes, Karagol~Ayan, Salimans, et~al.]{imagen}
Saharia, C., Chan, W., Saxena, S., Li, L., Whang, J., Denton, E.~L.,
  Ghasemipour, K., Gontijo~Lopes, R., Karagol~Ayan, B., Salimans, T., et~al.
\newblock Photorealistic text-to-image diffusion models with deep language
  understanding.
\newblock \emph{Advances in Neural Information Processing Systems},
  35:\penalty0 36479--36494, 2022.

\bibitem[Shahriari et~al.(2016)Shahriari, Swersky, Wang, Adams, and
  de~Freitas]{nando-bayesopt}
Shahriari, B., Swersky, K., Wang, Z., Adams, R.~P., and de~Freitas, N.
\newblock Taking the human out of the loop: A review of bayesian optimization.
\newblock \emph{Proceedings of the IEEE}, 104\penalty0 (1):\penalty0 148--175,
  2016.
\newblock \doi{10.1109/JPROC.2015.2494218}.

\bibitem[Snoek et~al.(2012)Snoek, Larochelle, and Adams]{snoek_bayesopt}
Snoek, J., Larochelle, H., and Adams, R.~P.
\newblock Practical bayesian optimization of machine learning algorithms.
\newblock \emph{Neural information processing systems}, 2012.

\bibitem[Sohl-Dickstein et~al.(2015)Sohl-Dickstein, Weiss, Maheswaranathan, and
  Ganguli]{diffusion-original}
Sohl-Dickstein, J., Weiss, E.~A., Maheswaranathan, N., and Ganguli, S.
\newblock Deep unsupervised learning using nonequilibrium thermodynamics.
\newblock In \emph{International Conference on Machine Learning}, 2015.

\bibitem[Song et~al.(2020)Song, Meng, and Ermon]{ddim}
Song, J., Meng, C., and Ermon, S.
\newblock Denoising diffusion implicit models.
\newblock \emph{arXiv preprint arXiv:2010.02502}, 2020.

\bibitem[Song et~al.(2021)Song, Sohl-Dickstein, Kingma, Kumar, Ermon, and
  Poole]{scorediff}
Song, Y., Sohl-Dickstein, J., Kingma, D.~P., Kumar, A., Ermon, S., and Poole,
  B.
\newblock Score-based generative modeling through stochastic differential
  equations.
\newblock In \emph{International Conference on Learning Representations}, 2021.

\bibitem[Srinivas et~al.(2010)Srinivas, Krause, Kakade, and
  Seeger]{ICML-2010-SrinivasKKS}
Srinivas, N., Krause, A., Kakade, S., and Seeger, M.~W.
\newblock {Gaussian Process Optimization in the Bandit Setting: No Regret and
  Experimental Design}.
\newblock In \emph{{Proceedings of the 27th International Conference on Machine
  Learning}}, pp.\  1015--1022. {Omnipress}, 2010.

\bibitem[Sutton et~al.(1999)Sutton, McAllester, Singh, and Mansour]{reinforce}
Sutton, R.~S., McAllester, D., Singh, S., and Mansour, Y.
\newblock Policy gradient methods for reinforcement learning with function
  approximation.
\newblock In \emph{Advances in Neural Information Processing Systems}, 1999.

\bibitem[Swaminathan \& Joachims(2015)Swaminathan and
  Joachims]{JMLR:v16:swaminathan15a_bandit}
Swaminathan, A. and Joachims, T.
\newblock Batch learning from logged bandit feedback through counterfactual
  risk minimization.
\newblock \emph{Journal of Machine Learning Research}, 16\penalty0
  (52):\penalty0 1731--1755, 2015.

\bibitem[Trabucco et~al.(2021)Trabucco, Kumar, Geng, and Levine]{coms}
Trabucco, B., Kumar, A., Geng, X., and Levine, S.
\newblock Conservative objective models for effective offline model-based
  optimization.
\newblock In \emph{International Conference on Machine Learning}, 2021.

\bibitem[Trabucco et~al.(2022)Trabucco, Geng, Kumar, and Levine]{design-bench}
Trabucco, B., Geng, X., Kumar, A., and Levine, S.
\newblock Design-bench: Benchmarks for data-driven offline model-based
  optimization.
\newblock In \emph{International Conference on Machine Learning}, pp.\
  21658--21676. PMLR, 2022.

\bibitem[Vincent(2011)]{dsm}
Vincent, P.
\newblock A connection between score matching and denoising autoencoders.
\newblock \emph{Neural computation}, 23\penalty0 (7):\penalty0 1661--1674,
  2011.

\bibitem[Zhang et~al.(2022)Zhang, Fu, Bengio, and Courville]{zhang2022unifying}
Zhang, D., Fu, J., Bengio, Y., and Courville, A.
\newblock Unifying likelihood-free inference with black-box sequence design and
  beyond.
\newblock In \emph{International Conference on Learning Representations}, 2022.
\newblock URL \url{https://openreview.net/forum?id=1HxTO6CTkz}.

\bibitem[Zheng et~al.(2022)Zheng, Zhang, and Grover]{onlinedt}
Zheng, Q., Zhang, A., and Grover, A.
\newblock Online decision transformer.
\newblock In \emph{International Conference on Machine Learning (ICML)}, pp.\
  27042--27059, 2022.

\bibitem[Zheng et~al.(2023)Zheng, Henaff, Amos, and Grover]{zheng2022semi}
Zheng, Q., Henaff, M., Amos, B., and Grover, A.
\newblock Semi-supervised offline reinforcement learning with action-free
  trajectories.
\newblock In \emph{International Conference on Machine Learning (ICML)}, 2023.

\bibitem[Zhu et~al.(2023)Zhu, Dang, and Grover]{zhuscaling}
Zhu, B., Dang, M., and Grover, A.
\newblock Scaling pareto-efficient decision making via offline multi-objective
  rl.
\newblock In \emph{International Conference on Learning Representations
  (ICLR)}, 2023.

\end{thebibliography}
\bibliographystyle{icml2023}

\newpage
\appendix
\onecolumn
\section{Notation and Experimental Details}
\label{app:nexp}
\subsection{Notation}
\begin{table}[ht!]
    \centering
    \begin{tabular}{ll}
    \multicolumn{1}{c}{\bf SYMBOL}  &\multicolumn{1}{c}{\bf MEANING}\\
    \hline\\
    $f$ & Black box function\\
    $\cX$ & Support of $f$\\
    $\xb^*$ & Optima (taken to be maxima for consistency)\\
    $\cD$ & Offline dataset \\
    $Q$ & Query budget for black-box function\\
    $K, \tau$ & Reweighting hyperparameters\\
    $B_i$ & Size of bin $i$
    \end{tabular}
    \caption{Important notation used in our paper}
    \label{tab:notation}
\end{table}

\subsection{Additional Experimental details}
\paragraph{Code} We build on top of \citet{scorediff2} implementation of score based diffusion models (linked \href{https://github.com/CW-Huang/sdeflow-light}{here}). We provide our code via an anonymized link \href{https://drive.google.com/drive/folders/1en5AUSsyftp_Qk9yYuV8_1sPVa19t2vD}{here}. All code we use is under the MIT licence.
\paragraph{Training details} We train our model on one RTX A5000 GPU and report results averaged over 5 seeds. For all tasks in Design-Bench~\citep{design-bench} we train with a batch size of $128$ for $1000$ epochs. For reweighting, we use a value of $K = 0.01 * N$ and $\tau = 0.1$. 
For discrete tasks, we follow a similar procedure to \citet{design-bench} and convert the $d$-to dimensional vector to a $c\times d$ size one hot vector. We then approximate logits by interpolating between a uniform distribution and the one hot distribution using a mixing factor of $0.6$. \sm{During training, we randomly set the conditioning value to 0 to jointly train a conditional and unconditional model with the same model. We use a dropout probability of 0.15, i.e. 15\% of the time the conditioning value is set to zero}

\paragraph{Evaluation} For all Design-bench tasks, we evaluate on a budget of $Q=256$ points. choice of the conditioning value is an important parameter for \name{}. As we see in Section \ref{sec:conditioning}, the predicted values increase upto around the dataset maxima before tapering off. This motivates us to use the dataset maxima for conditioning our model. Choosing the dataset maxima for conditioning has the advantages of being easy to implement for any type of dataset and not requiring any prior knowledge about the dataset (like the dataset maxima).
\paragraph{Design-Bench Tasks} We present additional information on the tasks we evaluate on in Design-Bench~\citep{design-bench}.
\begin{table}[!ht]
    \centering
\begin{tabular}{cccc}

\multicolumn{1}{c}{\bf TASK} & \multicolumn{1}{c}{\bf SIZE} & \multicolumn{1}{c}{\bf DIMENSIONS} & \multicolumn{1}{c}{\bf TASK MAX}\\
\hline\\
TFBind8          & 32898                          & 8  & 1.0\\ 
TFBind10         & 10000                          & 10  &   2.128    \\ 
ChEMBL         & 1093                          & 31     & 443000.0     \\
NAS         & 1771                          & 64         & 69.63    \\ 
D'Kitty & 10004                        & 56          & 340.0       \\ 
Ant    & 10004                        & 60        &      590.0     \\ 
Superconductor     & 17014                        & 86   & 185.0  \\ 

\end{tabular}
    \caption{Dataset Statistics}
    \label{tab:ds-stats}
\end{table}

\subsection{Unnormalized results}
We present the unnormalized results of Table \ref{tab:main_table} here, in Table \ref{tab:unnormalized}.
\begin{table}[!ht]
\centering
\resizebox{\columnwidth}{!}{
\begin{tabular}{ccccccc}

\multicolumn{1}{c}{\textbf{BASELINE}}  &\multicolumn{1}{c}{\textbf{TFBIND8}} &\multicolumn{1}{c}{\textbf{TFBIND10}} &\multicolumn{1}{c}{\textbf{SUPERCON.}} &\multicolumn{1}{c}{\textbf{ANT}}  &\multicolumn{1}{c}{\textbf{D'KITTY}} &\multicolumn{1}{c}{\textbf{CHEMBL}} \\
\hline\\
$\cD$ (best) & $0.439$ & $0.00532$ & $74.0$ & $165.326$ & $199.231$ & $443000.000$\\
\hline\\
CbAS                          &  $0.958 \pm 0.018$ &  $0.761 \pm 0.067$ &  $83.178 \pm 15.372$ &    $468.711 \pm 14.593$ &    $213.917 \pm 19.863$ & $389000.000\pm 500.000$\\
GP-qEI                        &  $0.824 \pm 0.086$ &  $0.675 \pm 0.043$ &  ${92.686 \pm 3.944}$ &     $480.049 \pm 0.000$ &     $213.816 \pm 0.000$ & $387950.000\pm 0.000$\\
CMA-ES                        &  $0.933 \pm 0.035$ &  ${0.848 \pm 0.136}$ & $90.821 \pm 0.661$ &  ${1016.409 \pm 906.407}$ &       $4.700 \pm 2.230$ & $388400.000\pm 400.000$\\
Gradient Ascent               &  ${0.981 \pm 0.010}$ &  $0.770 \pm 0.154$ &  ${93.252 \pm 0.886}$ &    $-54.955 \pm 33.482$ &    $226.491 \pm 21.120$ & $390050.000 \pm 2000.000$\\
REINFORCE                     &  $0.959 \pm 0.013$ &  $0.692 \pm 0.113$ &  $89.027 \pm 3.093$ &   $-131.907 \pm 41.003$ &  $-301.866 \pm 246.284$ & $388400.000\pm 2100.000$\\
MINs                          &  $0.938 \pm 0.047$ &  $ 0.770 \pm 0.177$ & $89.469 \pm 3.227$ &    $533.636 \pm 17.938$ &    $272.675 \pm 11.069$ & ${390950.000 \pm 200.000}$\\
COMs                          &  $0.964 \pm 0.020$ &  $0.750 \pm 0.078$ & $78.178 \pm 6.179$ &    $540.603 \pm 20.205$ &     ${277.888 \pm 7.799}$ & $390200.000\pm 500.000$3\\
\hline\\
\name{} & $0.971\pm 0.005$ & $0.885\pm 0.367$ & $103.600\pm 8.139$ & $548.227\pm 11.725$ & $250.529\pm 10.992$ & $387950.000\pm 1050.000$\\
\end{tabular}
}
\caption{Unnormalized counterpart to results presented in Table \ref{tab:main_table}}
\label{tab:unnormalized}
\end{table}

\subsection{HopperController}
\label{sec:no_hopper}
We don't include the Hopper task in our results in Table \ref{tab:main_table} because of inconsistencies between the offline dataset values and the values obtained when running the oracle. A similar issue was noticed by \citet{bonet} in the Hopper task, and it seems to be a known bug in Design-bench (see \href{https://github.com/brandontrabucco/design-bench/issues/8}{here}). Due to such discrepancies, we decided not to include Hopper in our analysis.

\begin{figure}[!ht]
\centering
\begin{minipage}[t]{.49\textwidth}
    \centering
    \includegraphics[scale=0.45]{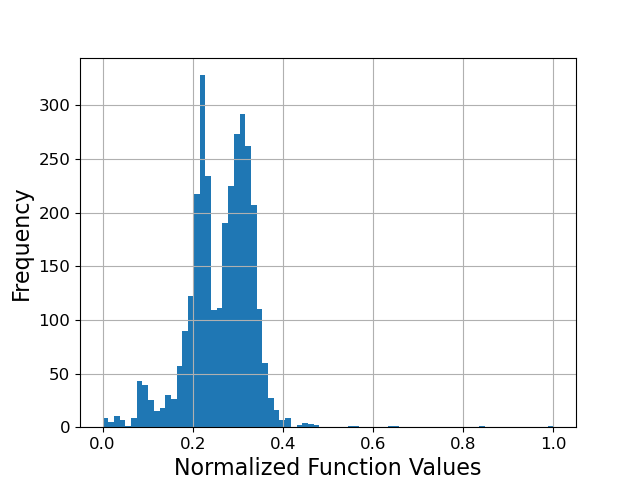}
    \captionof{figure}{Histogram of normalized function values in the Hopper dataset. The distribution is highly skewed towards low function values. Plots taken from \citet{bonet}}
    \label{fig:hopper1}
\end{minipage}%
\hfill
\begin{minipage}[t]{.49\textwidth}
    \centering
    \includegraphics[scale=0.42]{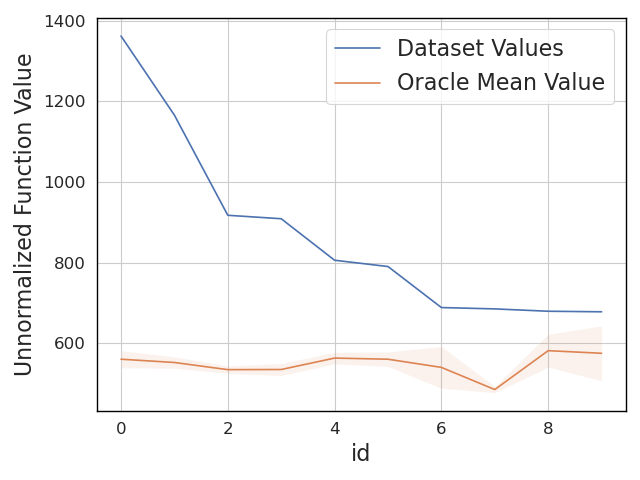}
    \captionof{figure}{Dataset values vs Oracle values for top $10$ points. Oracle being noisy, we show mean and standard deviation over $20$ runs. Plots taken from \citet{bonet}.}
    \label{fig:hopper2}
\end{minipage}
\end{figure}

\section{Additional Ablations and Analysis}
\label{app:ablations}
\subsection{Effect of evaluation budget}
\begin{figure}[ht!]
    \centering
    \begin{subfigure}{0.5\textwidth}
        \includegraphics[width=\linewidth]{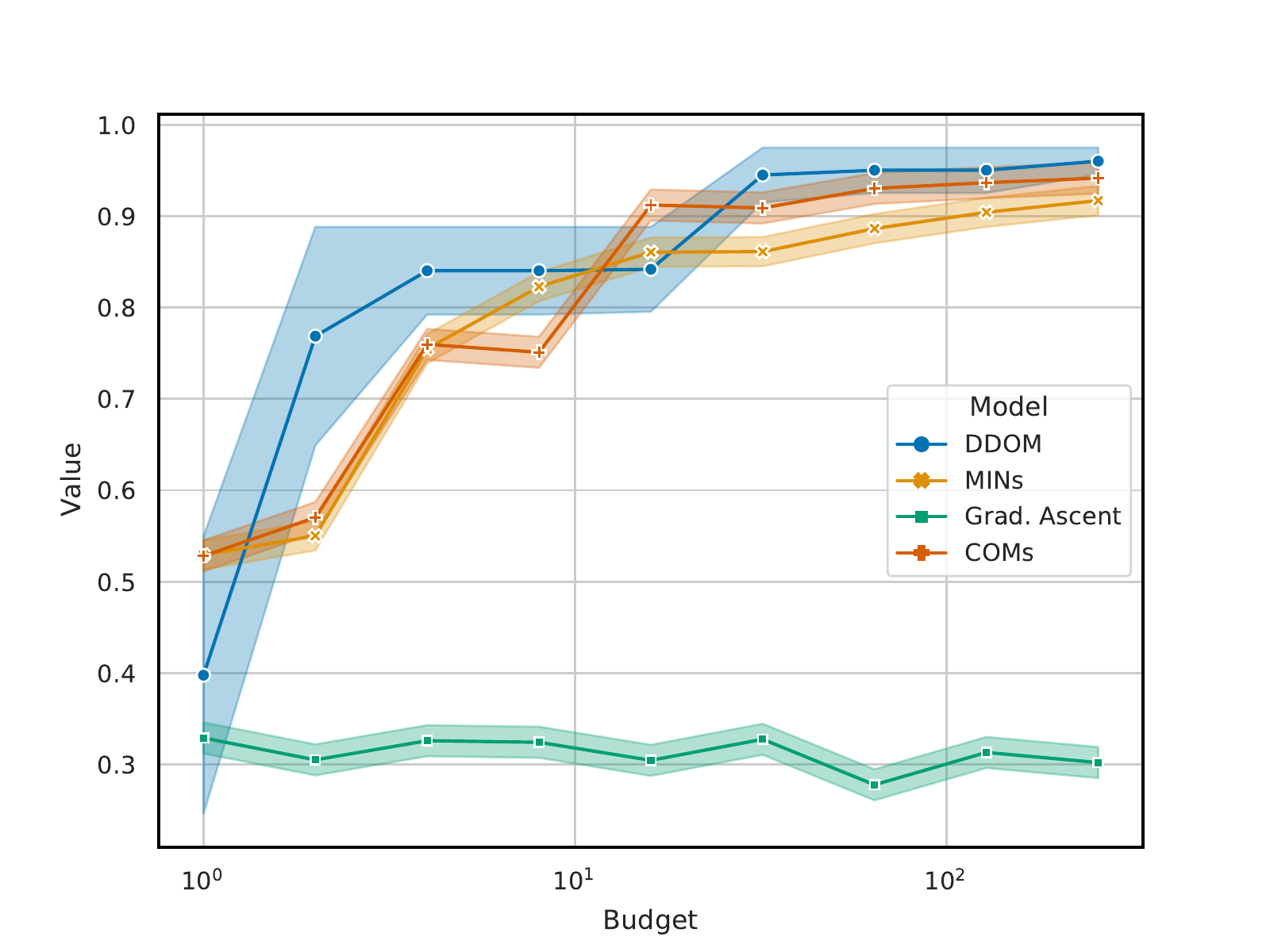}
    \end{subfigure}
    \caption{Plot of maximum objective value versus budget $Q$ for Ant, averaged over 5 seeds. We find that \name{} outperforms MINs, COMs and Grad. Ascent on most choices of budget, indicating that \name{} is consistently able to find good points.}
    \label{fig:budget}
\end{figure}

In this experiment, we vary the evaluation budget $Q$ on Ant for \name{} to see how sensitive the quality of predictions are to the budget $Q$. The results are shown in Figure \ref{fig:budget}. We see that \name{} outperforms multiple baselines (namely MINs, COMs and Grad. Ascent) for various budget values, indicating that \name{} is consistently able to find good points.

\subsection{Ablation on reweighting parameters}
During reweighting, we reweight the loss using the formula 
\begin{equation}
    w_i = \frac{|B_i|}{|B_i| + K} \exp{\left ( \frac{-|\hat{y} - y_{b_i}|}{\tau} \right )}
\end{equation}
In the equation, $K$ and $\tau$ act as smoothing parameters. $\tau$ controls how much weight is given to good bins (bins with good points) versus bad bins. Lowering $\tau$ would lead to lower weights for low quality bins (bins with low scores) and vice versa.
A smaller value of $K$ leads to bin size being an irrelevant factor in reweighting, meaning that weights are computed solely according to the values of the points. A large value of $K$ would lead to the reweighting scaling roughly linearly with the size of the bin.

\begin{figure}[ht!]
    \centering
    \begin{subfigure}{0.4\textwidth}
        \includegraphics[width=\linewidth]{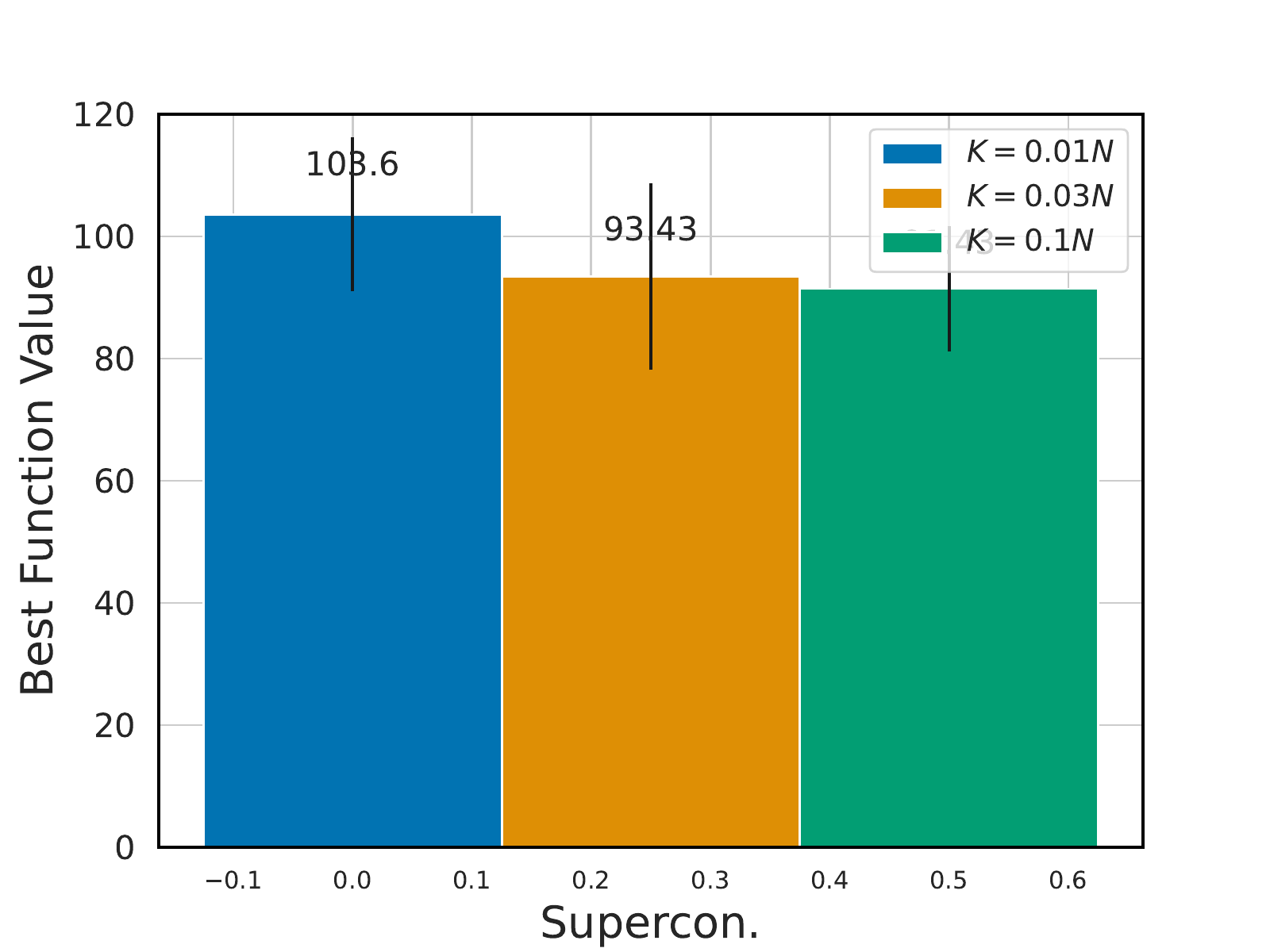}
        
    \end{subfigure}
    \begin{subfigure}{0.4\textwidth}
        \includegraphics[width=\linewidth]{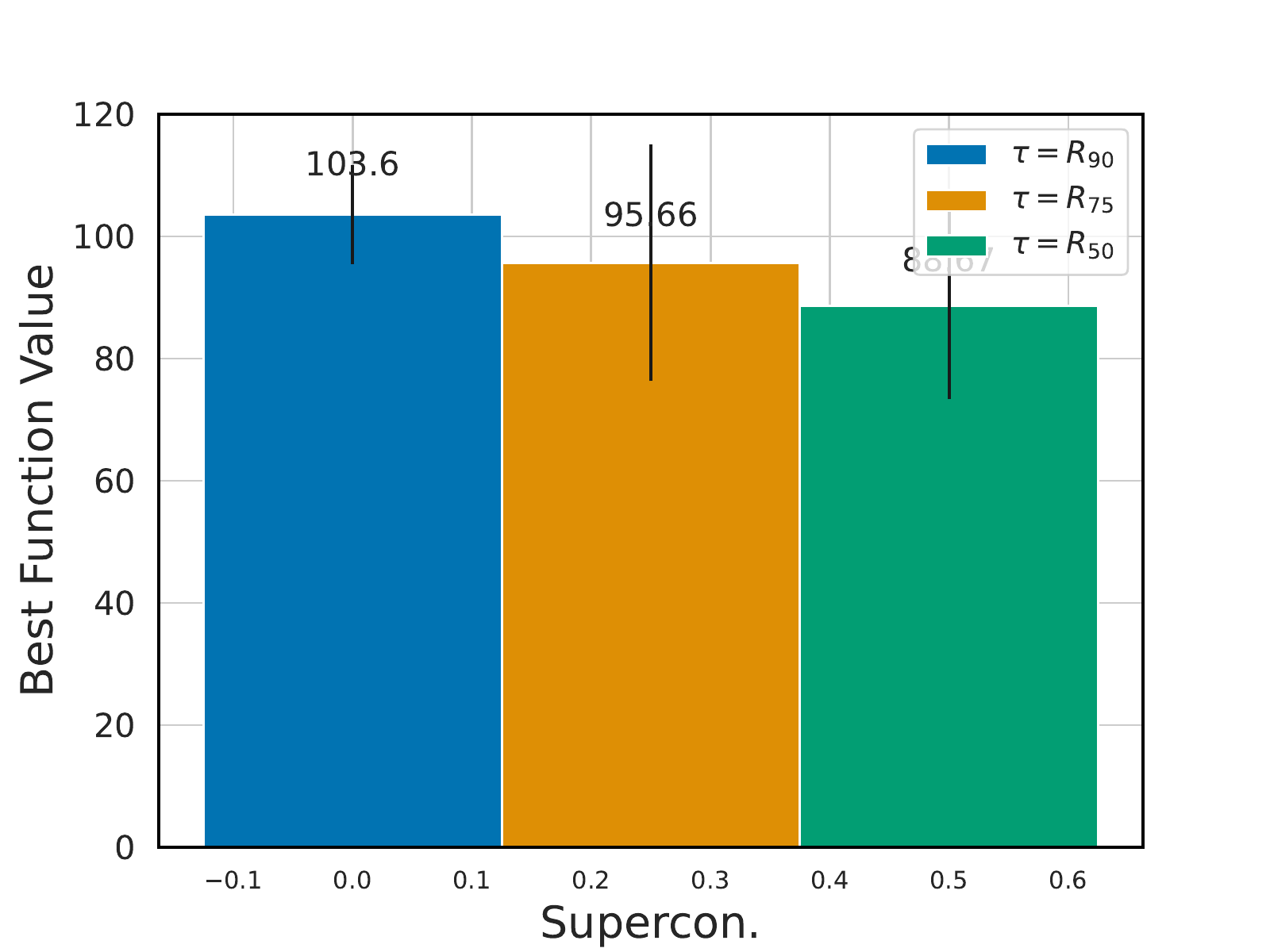}
    \end{subfigure}
    \caption{Plot of function value of Superconductor for various values of $K$ and $\tau$}
    \label{fig:ktau}
\end{figure}

In Figure \ref{fig:ktau} we plot the function value of superconductor achieved when using various values of $K$ and $\tau$.

\sm{We further run an ablation on the number of bins on AntMorphology where we try out 3 different values of $N_B$. The results are in Table \ref{tab:nb}. We find that for 1 bin, we see a significant difference, in performance, but the difference is much smaller between 32 and 64 bins.}

\begin{table}[!ht]
\centering
\begin{tabular}{cc}
$N_B$ & \textbf{SCORE}\\
\hline\\
$N_B = 1$\% & $480.820$\\
$N_B = 32$\% & $541.640$\\
$N_B = 64$\% & $548.227$\\
\end{tabular}
\caption{Ablation on number of bins}
\label{tab:nb}
\end{table}

\subsection{Visualizing the Design-Bench datasets}
\begin{figure}[ht!]
     \centering
          \begin{subfigure}{0.4\textwidth}
         \centering
         \includegraphics[width=\textwidth]{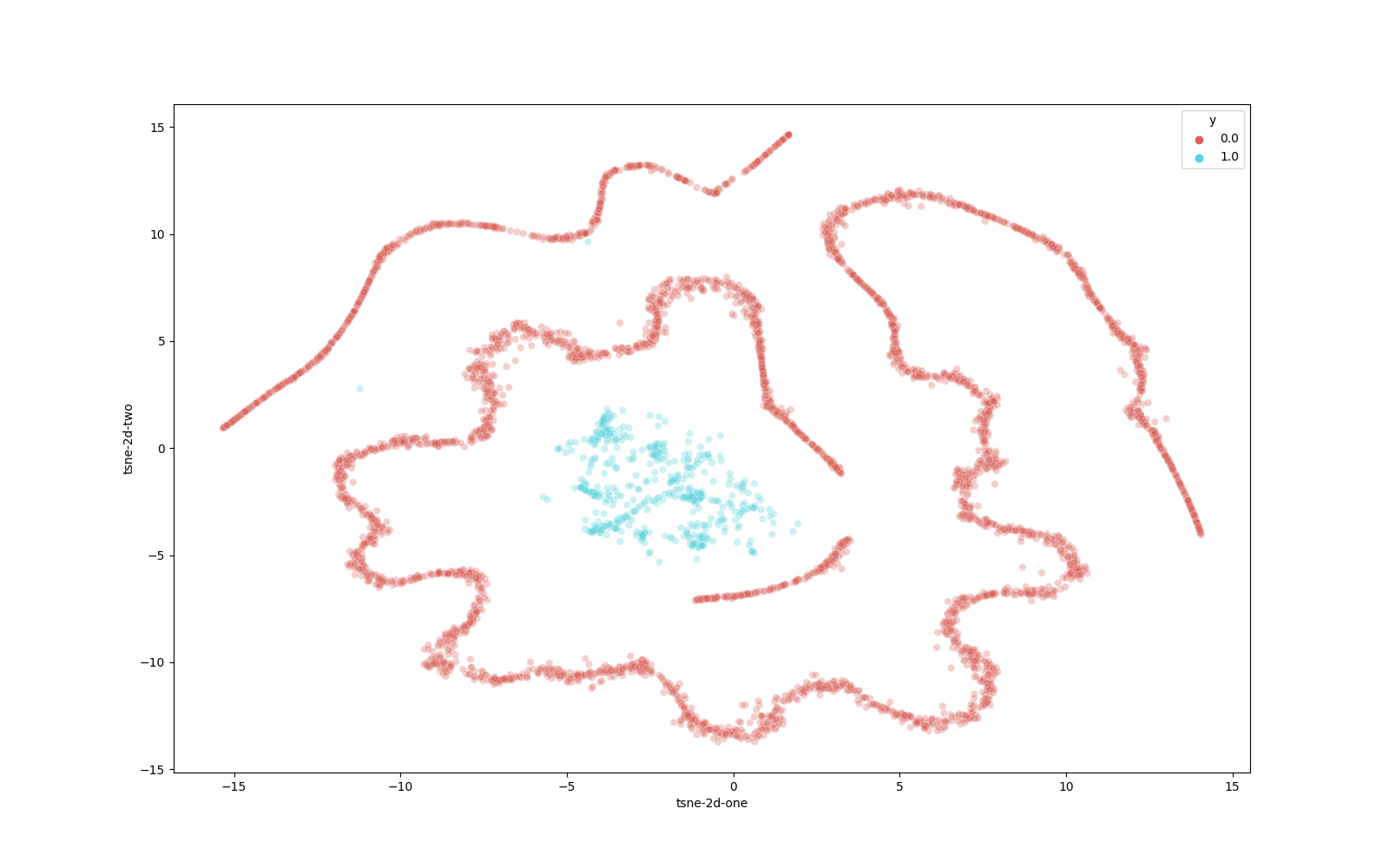}
         \caption{Ant}
    \end{subfigure}
     \begin{subfigure}{0.4\textwidth}
         \centering
         \includegraphics[width=\textwidth]{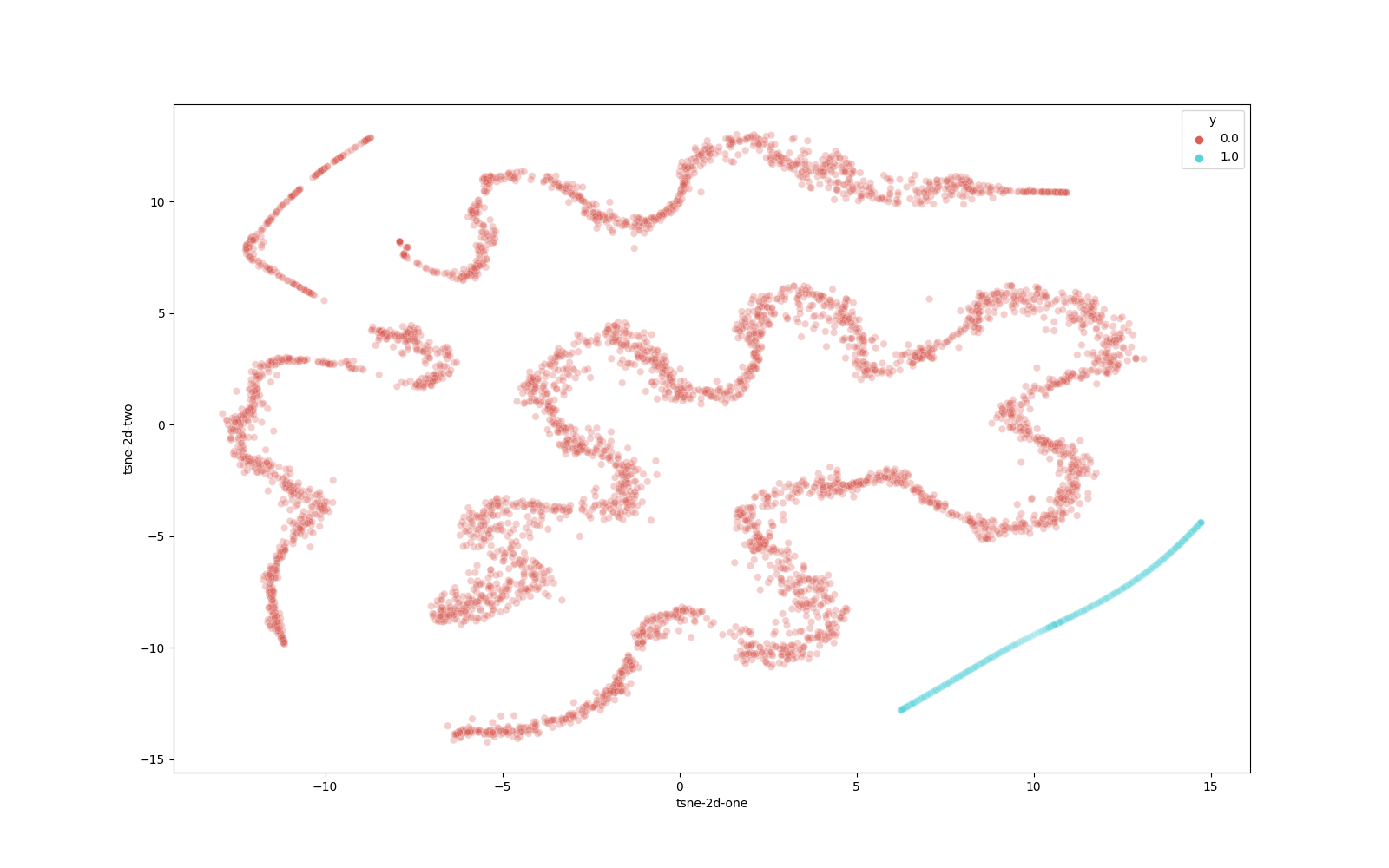}
         \caption{DKitty}
        \end{subfigure}
     \begin{subfigure}{0.4\textwidth}
         \centering
         \includegraphics[width=\textwidth]{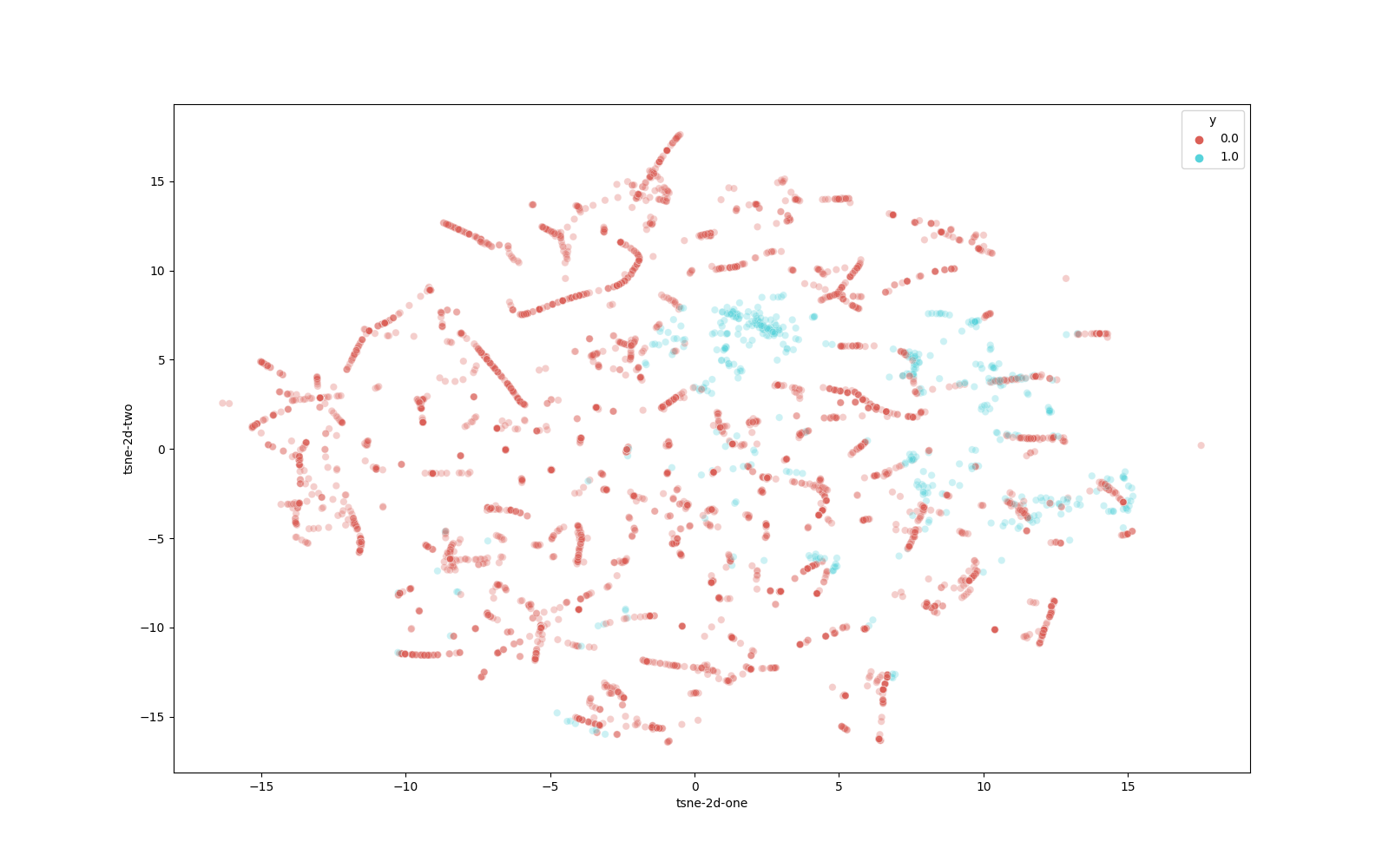}
         \caption{Superconductor}
        \end{subfigure}
    \caption{t-SNE plots of dataset and predicted points for D'Kitty, Ant and Superconductor. The red points refer to dataset points, and the blue points refer to the points predicted by our model. We find that the points predicted by our model do not just follow the contours of the dataset points, indicating that the model is just not memorizing the dataset}
    \label{fig:tsne}
    \end{figure}

In Figure \ref{fig:tsne}, we show plots of t-SNE of the dataset and predicted points for 3 continuous tasks, D'Kitty, Ant and Superconductor. The red points refer to dataset points, and the blue points refer to the points predicted by our model. We find that the points predicted by our model do not just follow the contours of the dataset points, indicating that the model is just not memorizing the dataset.

\subsection{Plotting the contours of the score function}
\begin{figure}[ht!]
    \centering
    \begin{subfigure}[b]{0.15\textwidth}
    \includegraphics[width=\linewidth]{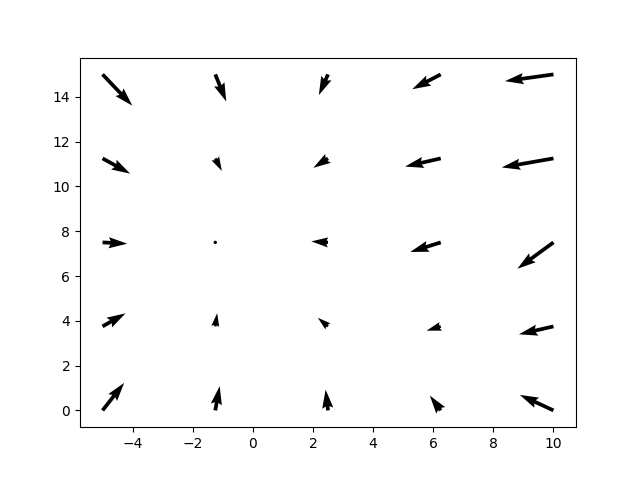}
    \caption{$t = 0$}
    \end{subfigure}
    \begin{subfigure}[b]{0.15\textwidth}
    \includegraphics[width=\linewidth]{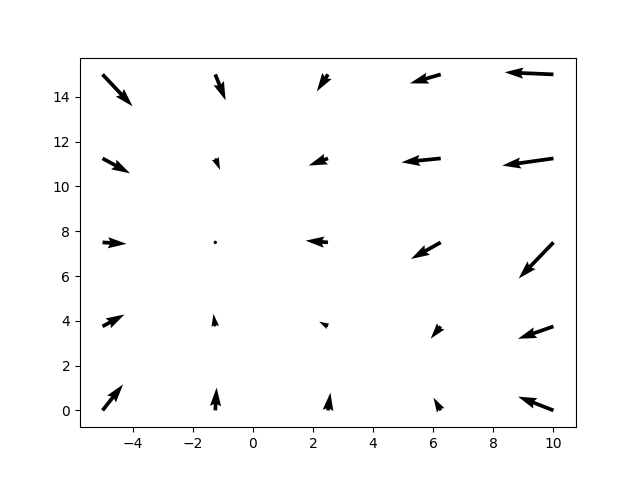}
    \caption{$t = 0.125$}
    \end{subfigure}
    \begin{subfigure}[b]{0.15\textwidth}
    \includegraphics[width=\linewidth]{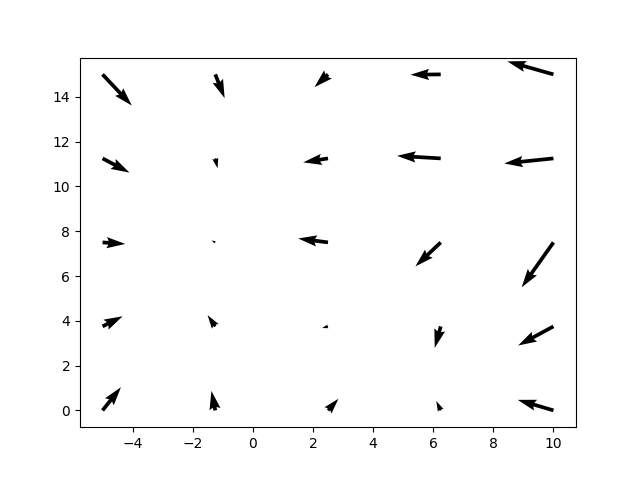}
    \caption{$t=0.25$}
    \end{subfigure}
    \begin{subfigure}[b]{0.15\textwidth}
    \includegraphics[width=\linewidth]{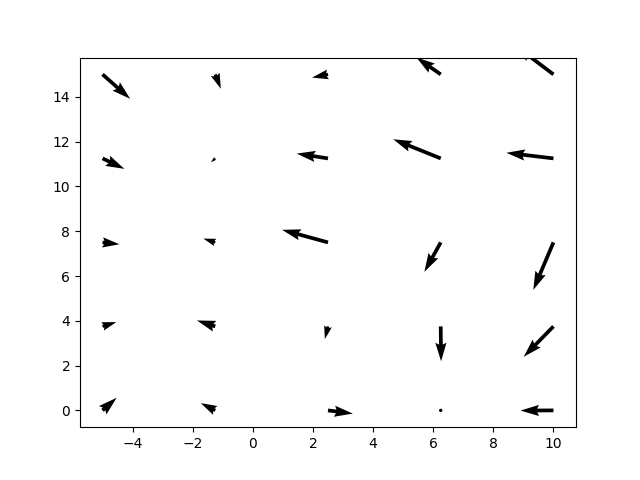}
    \caption{$t=0.5$}
    \end{subfigure}
    \begin{subfigure}[b]{0.15\textwidth}
    \includegraphics[width=\linewidth]{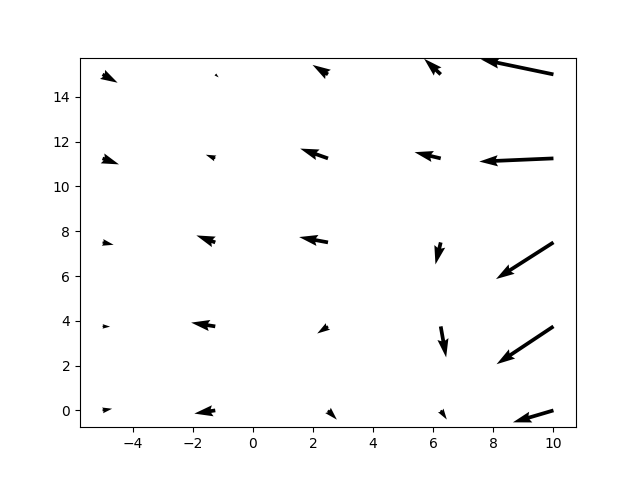}
    \caption{$t=0.875$}
    \end{subfigure}
    \begin{subfigure}[b]{0.15\textwidth}
    \includegraphics[width=\linewidth]{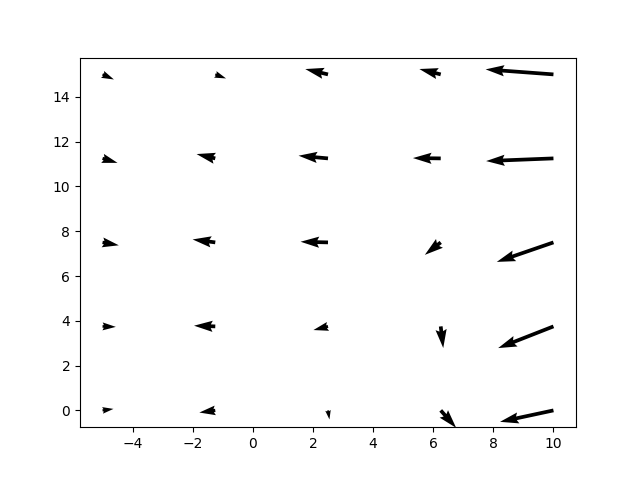}
    \caption{$t=1.0$}
    \end{subfigure}
    
    \caption{In this plot, we show the the contours of the score function on the Branin task across various timesteps. This illustrates how the learned score function varies with time.}
    \label{fig:scorefunc}
\end{figure}

In Figure \ref{fig:scorefunc}, we show the the contours of the score function on the Branin task across various timesteps. This illustrates how the learned score function varies with time.

\subsection{Effect of adding randomness to data}
\sm{We run an experiment where we add noise to the function evaluations of Superconductor. The results are in Table \ref{tab:random_noise}. We find that with only a little noise, the performance isn’t affected too much. Even a significant fraction like 20\% still sees only a moderate decline in performance.}

\begin{table}[!ht]
\centering
\begin{tabular}{cc}
\textbf{NOISE (as a \% of $\cD$ (best))} & \textbf{SCORE}\\
\hline\\
$0$\% & $103.600$\\
$2$\% & $98.470$\\
$20$\% & $91.252$\\
$100$\% & $79.374$\\
\end{tabular}
\caption{Ablation on adding additional noise to data. Noise added is mentioned as a \% of the dataset maxima}
\label{tab:random_noise}
\end{table}

\subsection{Effect of size of the offline dataset}

\sm{We run an experiment on Superconductor where we randomly withhold some percentage of points from the offline dataset. The results are in Table \ref{tab:smalldata} We find that even when removing parts of the offline dataset, the model is able to perform better than the dataset optima, indicating that there is a degree of generalizability.}

\begin{table}[!ht]
\centering
\begin{tabular}{cc}
\textbf{\% WITHHELD} & \textbf{SCORE}\\
\hline\\
$10$\% & $102.340$\\
$50$\% & $97.200$\\
$90$\% & $89.240$\\
$99$\% & $83.120$\\
\end{tabular}
\caption{Ablation on size of offline data}
\label{tab:smalldata}
\end{table}

\section{Societal Impact}
\label{app::impact}
While we don't anticipate anything to be inherently malicious about our work, it is possible that our method (and other optimizers like it) could be used in harmful settings (e.g. optimizing for drugs with adverse side effects). This is something to be careful of when deploying such optimizing algorithms in the real world.

\end{document}